\documentclass[letterpaper, 10 pt, conference]{ieeeconf}  % Comment this line out if you need a4paper

\IEEEoverridecommandlockouts                              % This command is only needed if 
\usepackage{amsmath} % assumes amsmath package installed
\usepackage{amssymb}  % assumes amsmath package installed

\let\labelindent\relax  % Necessary to include [inline]enumitem
\usepackage{graphicx}
\usepackage{subcaption}
\usepackage[hidelinks]{hyperref}
\usepackage{comment}  % TODO delete
\usepackage{calc}
\usepackage{tikz}
\usetikzlibrary{arrows,shapes,positioning}
\usepackage{amsfonts}
\usepackage{algorithm, algpseudocode}
\usepackage{bm}
\usepackage{multirow}
\usepackage{siunitx}
\usepackage[inline]{enumitem}
\usepackage{xspace}
\newcommand*{\eg}{e.g.\@\xspace}
\newcommand*{\ie}{i.e.\@\xspace}

\newcommand*{\wrt}{w.r.t.\@\xspace}
\newcommand*{\vs}{vs.\@\xspace}

\definecolor{blue}{RGB}{83,125,221}
\definecolor{red}{RGB}{218,76,76}
\definecolor{green}{RGB}{71,154,95}
\definecolor{purple}{RGB}{125,84,178}
\definecolor{grey}{RGB}{161,169,173}

\newcommand{\defineleg}[2]{
    \newcommand{#1}{\raisebox{2pt}{\tikz{\draw[#2, solid, line width=0.9pt](0,0)--(6mm,0);}}}
}

\defineleg{\legblue}{blue}
\defineleg{\legred}{red}
\defineleg{\leggreen}{green}
\defineleg{\legpurple}{purple}
\defineleg{\leggrey}{grey}

\title{\LARGE \bf
Benchmarking Population-Based Reinforcement Learning across Robotic Tasks with GPU-Accelerated Simulation
}

\author{Asad Ali Shahid$^{1}$, Yashraj Narang$^{2}$, Vincenzo Petrone$^{1,3}$, Enrico Ferrentino$^3$,\\
Ankur Handa$^2$, Dieter Fox$^2$, Marco Pavone$^4$ and Loris Roveda$^{1,4*}$% <-this % stops a space
\thanks{$^{1}$ Dalle Molle Institute for Artificial Intelligence, IDSIA USI-SUPSI}%
\thanks{$^2$ NVIDIA Corporation, Santa Clara (CA), USA}%
\thanks{$^3$ University of Salerno, Fisciano (SA), Italy}%
\thanks{$^4$ Stanford University, Stanford (CA), USA}%
\thanks{$^{*}$ corresponding author {\tt\small loris.roveda@idsia.ch}}%
\thanks{Accepted version, published at \url{https://doi.org/10.1109/CASE58245.2025.11163870} -- 2025 IEEE 21st International Conference on Automation Science and Engineering (CASE)}
}

\begin{document}

\maketitle
\thispagestyle{empty}
\pagestyle{empty}

%%%%%%%%%%%%%%%%%%%%%%%%%%%%%%%%%%%%%%%%%%%%%%%%%%%%%%%%%%%%%%%%%%%%%%%%%%%%%%%%
\begin{abstract}  % At most 1800 characters
In recent years, deep reinforcement learning (RL) has shown its effectiveness in solving complex continuous control tasks.
However, this comes at the cost of an enormous amount of experience required for training, exacerbated by the sensitivity of learning efficiency and the policy performance to hyperparameter selection, which often requires numerous trials of time-consuming experiments.
This work leverages a Population-Based Reinforcement Learning (PBRL) approach and a GPU-accelerated physics simulator to enhance the exploration capabilities of RL by concurrently training multiple policies in parallel.
The PBRL framework is benchmarked against three state-of-the-art RL algorithms -- PPO, SAC, and DDPG -- dynamically adjusting hyperparameters based on the performance of learning agents.
The experiments are performed on four challenging tasks in Isaac Gym -- \textit{Anymal Terrain}, \textit{Shadow Hand}, \textit{Humanoid}, \textit{Franka Nut Pick} -- by analyzing the effect of population size and mutation mechanisms for hyperparameters.
The results show that PBRL agents achieve superior performance, in terms of cumulative reward, compared to non-evolutionary baseline agents. 
Moreover, the trained agents are finally deployed in the real world for a \textit{Franka Nut Pick} task.
To our knowledge, this is the first sim-to-real attempt for deploying PBRL agents on real hardware.
Code and videos of the learned policies are available on our \href{https://sites.google.com/view/pbrl}{project website}.
\end{abstract}

\begin{keywords}
Reinforcement Learning, Evolutionary Learning, Sim-to-Real, Dexterous Manipulation
\end{keywords}

\section{Introduction}

\begin{figure}[t!]
\centering
    \begin{subfigure}{0.49\columnwidth}
        \includegraphics[width=\columnwidth]{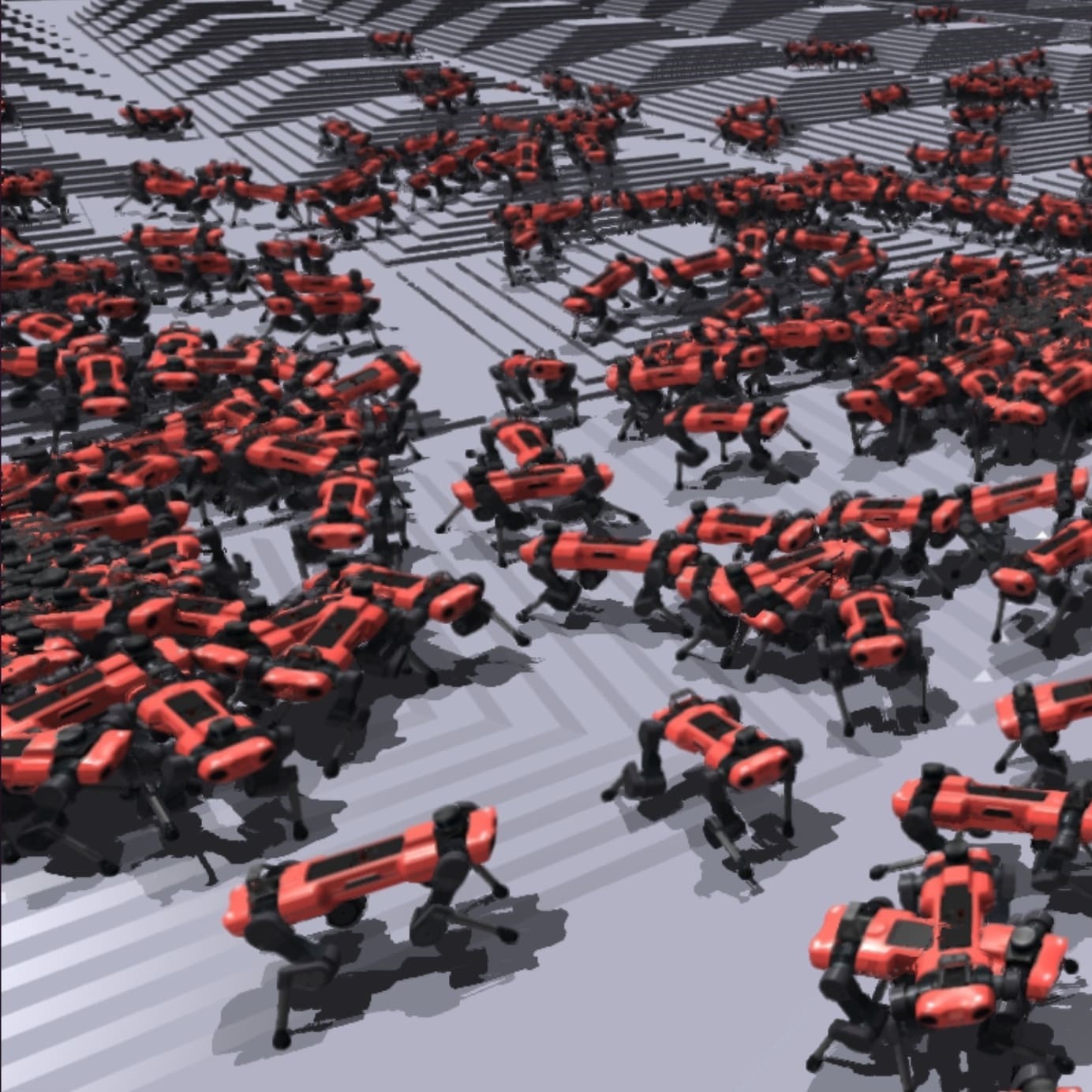}
        \caption{\textit{Anymal Terrain}}
        \label{fig:anymal}
    \end{subfigure}
    \hfill
    \begin{subfigure}{0.49\columnwidth}
        \includegraphics[width=\columnwidth]{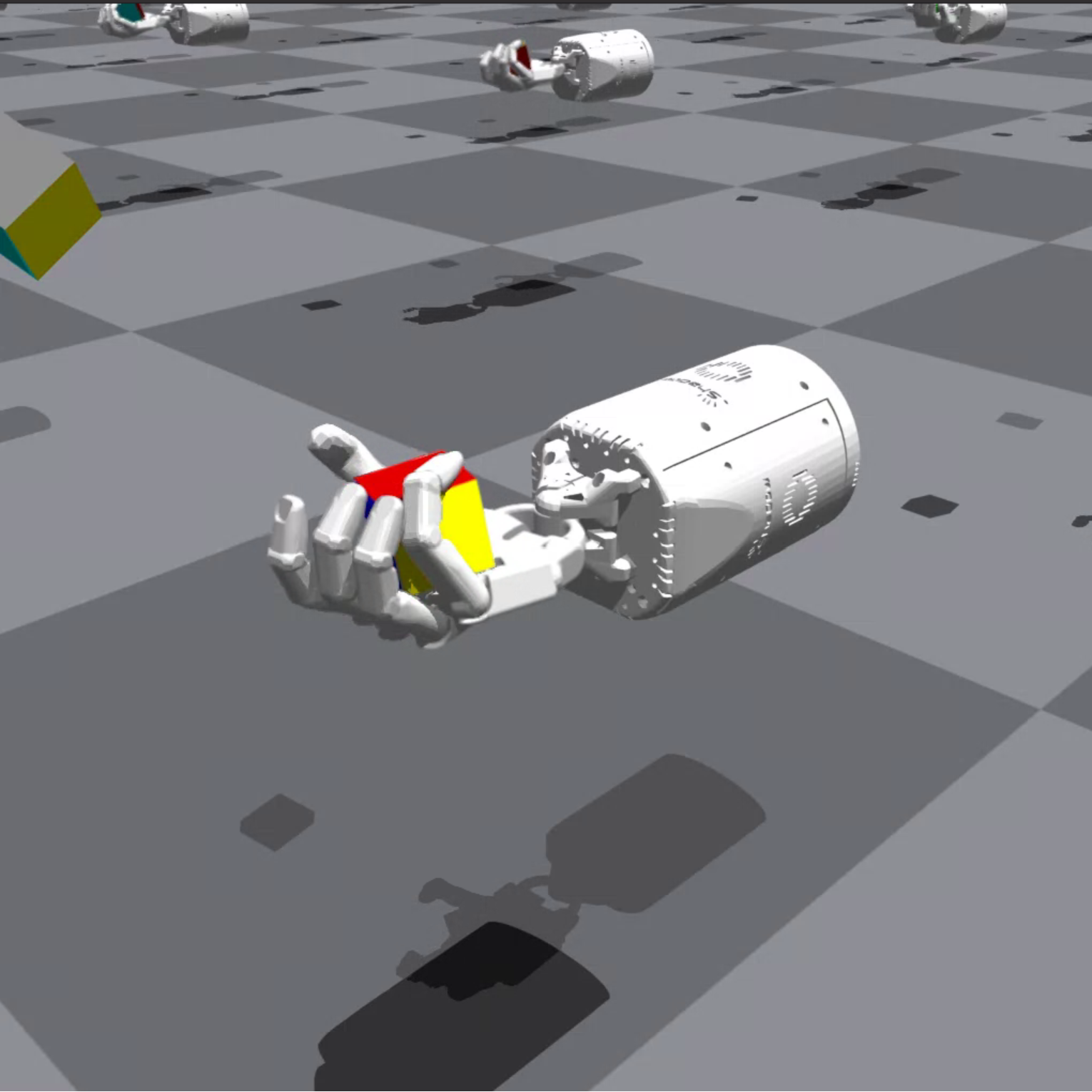}
        \caption{\textit{Shadow Hand}}
        \label{fig:shadow-hand}
    \end{subfigure}
    \\
    \begin{subfigure}{0.49\columnwidth}
        \includegraphics[width=\columnwidth]{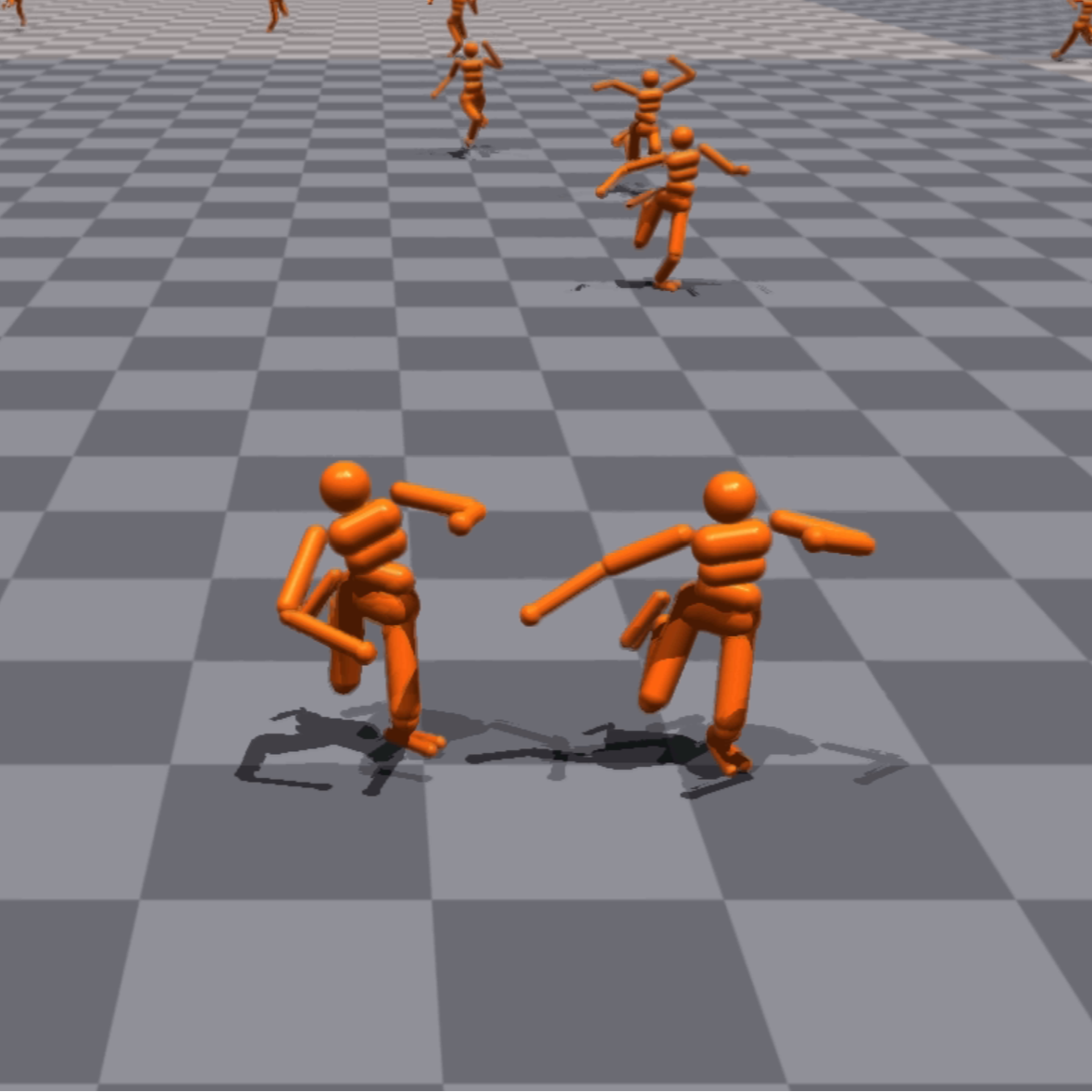}
        \caption{\textit{Humanoid}}
        \label{fig:humanoid}
    \end{subfigure}
    \hfill
    \begin{subfigure}{0.49\columnwidth}
        \includegraphics[width=\columnwidth]{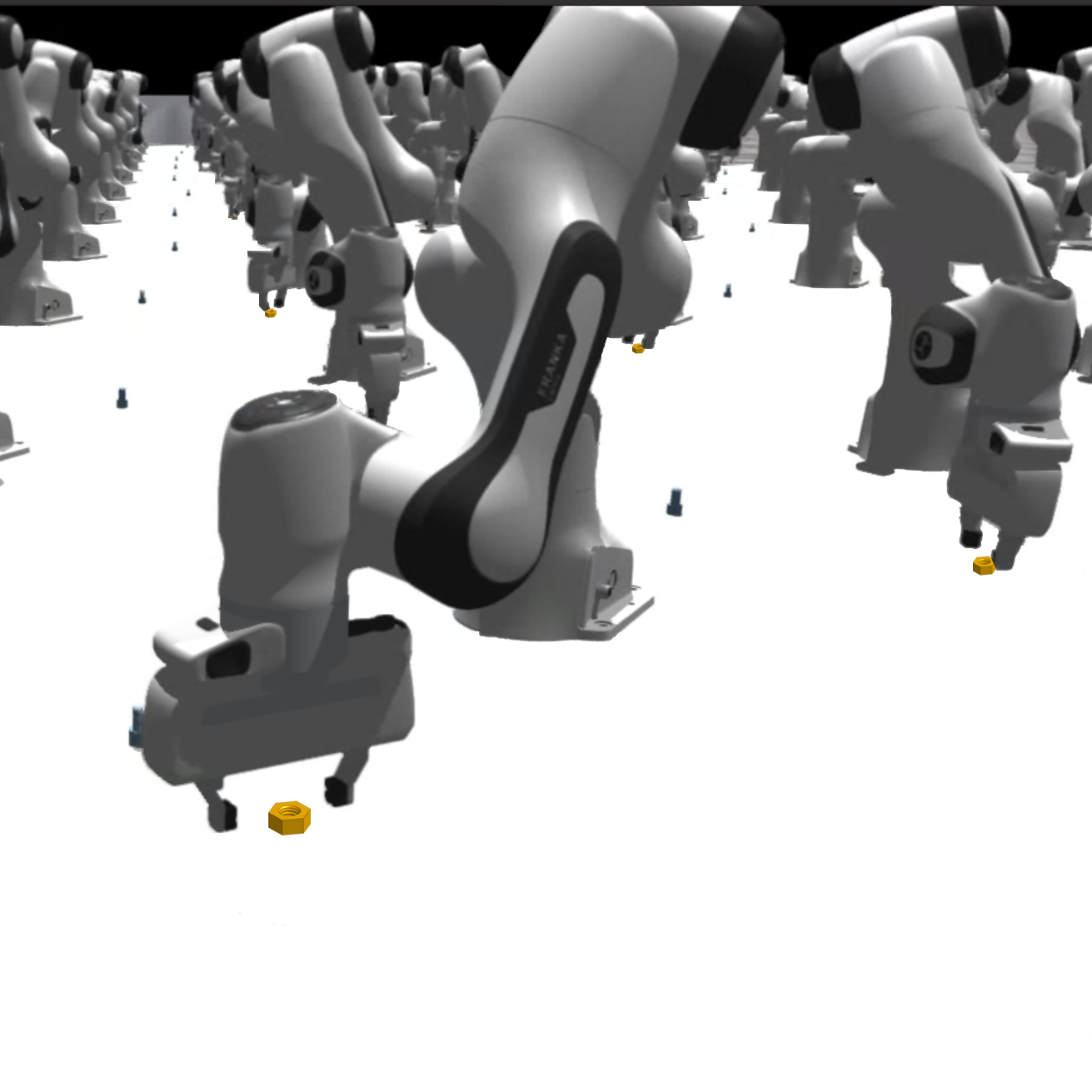}
        \caption{\textit{Franka Nut Pick}}
        \label{fig:franka-nut-pick}
    \end{subfigure}
    \caption{Simulated experiments are performed on four Isaac Gym benchmark tasks: (\subref{fig:anymal}) \textit{Anymal Terrain}, to teach a quadruped robot to navigate uneven terrain; (\subref{fig:shadow-hand}) \textit{Shadow Hand}, to re-orient cube to a desired configuration with a robot hand; (\subref{fig:humanoid}) \textit{Humanoid}, for bipedal locomotion; and (\subref{fig:franka-nut-pick}) \textit{Franka Nut Pick}, to grasp and lift a nut from a surface.}
    \label{fig:envs}
    \vspace*{-1em}
\end{figure}

Many domains have seen tremendous advancements of reinforcement learning (RL) applications in recent years, ranging from playing challenging games \cite{silver2018general, berner2019dota} to learning high-dimensional continuous control in robotics \cite{shahid2021decentralized, narang2022factory, shahid2022continuous}.
Tasks such as dexterous manipulation \cite{andrychowicz2020learning}, legged locomotion \cite{margolis2022rapid}, and mobile navigation \cite{kahn2018self} have been learned using deep RL. 
A primary challenge in training RL policies is the need for large amounts of training data.
RL methods rely on effective exploration to discover control policies, which can be particularly challenging when operating in high-dimensional continuous spaces \cite{Xu2022}.
Moreover, the performance of the learned policy is highly dependent on the tedious hyperparameters tuning procedure, a very time-consuming process often requiring several manual trials to determine the best values for the specific task and the learning environment.

One way to deal with the problem of data inefficiency is to train in simulation before transferring to reality \cite{tang2023industreal, akkaya2019solving, miki2022learning}.
However, the time required to train the policy in simulation increases significantly with the task complexity.
For example, in \cite{akkaya2019solving}, learning a block re-orientation task with a robot hand took around 14 days and enormous computing resources.
In addition, policies trained in simulation often fail to perform on a real system due to discrepancies between the simulation and the real world.
Recent advances in GPU-accelerated simulation, such as Isaac Gym \cite{makoviychuk2021isaac, handa2022dextreme}, have made it possible to run thousands of parallel environments on a single GPU, which significantly reduces the training times. 
Nevertheless, successfully training RL policies still requires carefully tuned hyperparameters to explore efficiently.

\subsection{Related Works}

\subsubsection{Massively-Parallel Simulation}

The advent of GPU-based simulation has significantly improved simulation throughput by enabling massive parallelism on a single GPU \cite{makoviychuk2021isaac, liang2018gpu}.
A number of recent works have exploited this parallelism to demonstrate impressive performance on challenging control problems using RL \cite{handa2022dextreme, allshire2022transferring, rudin2022learning}.
Almost all recent works use the same algorithm, \ie Proximal Policy Optimization (PPO) \cite{schulman2017proximal} to train RL policies; other common approaches include off-policy techniques, \eg Soft Actor-Critic (SAC) \cite{haarnoja2018soft} and Deep Deterministic Policy Gradient (DDPG) \cite{lillicrap2015continuous}.
While simple and effective, all these algorithms require a range of hyperparameters that need to be tuned for each task to ensure sufficient exploration and stabilize training.

\subsubsection{Population-Based RL}

Population-based approaches provide a promising solution for exploration and hyperparameter tuning by training a set of policies instead of independent ones.
By leveraging multiple agents, these methods improve robustness and stabilize training through dynamic hyperparameters adaptation.
Prior research has demonstrated the effectiveness of such approaches in training deep RL policies, particularly in strategy games and multi-agent interaction \cite{vinyals2019grandmaster, jaderberg2018human, flajolet2022fast}.
However, the application of population-based RL (PBRL) methods to robotics remains largely unexplored.
This limitation arises primarily due to the computational demands and increased training time, which scale linearly with the number of agents when using CPU-based simulators like MuJoCo \cite{todorov2012mujoco}.
Efficient data collection in such settings typically requires multiple worker machines running separate simulation instances.
In contrast, Isaac Gym enables large-scale parallel simulation of thousands of robots, significantly accelerating data generation and making it well-suited for training PBRL agents \cite{makoviychuk2021isaac}.

A key advantage of training multiple RL agents is its potential for meta-optimization by combining learning and evolutionary mechanisms \cite{ackley1991interactions}.
One widely adopted PBRL method is population-based training (PBT) \cite{jaderberg2017population}, where multiple policies are trained concurrently to enhance exploration and promote diverse behaviors.
In PBT, an inner loop optimizes policies, while an outer evolutionary loop periodically selects and mutates the best-performing agents.
More recently, DexPBT \cite{petrenko2023dexpbt}, a decentralized PBRL approach, has demonstrated impressive results in dexterous manipulation using parallel simulations.
By distributing evolutionary updates across computing nodes, the authors successfully evolved on-policy RL agents.
However, no \textit{sim-to-real} transfer was conducted, underlining the challenges of deploying these policies on physical robots.

In this paper, we exploit parallel GPU simulation to benchmark PBRL algorithms across 3 RL baselines, including both on-policy (PPO) and off-policy (SAC, DDPG) methods on 4 distinct robotic environments, also evaluating the impact of different hyperparameter mutation mechanisms.
Furthermore, to increase the significance and extend the applicability of the results, we transfer policies to a real robot, without an additional learning adaptation phase.

\subsubsection{Sim-to-Real Transfer}

Despite the calibration efforts to model the physical system accurately, simulation is still a rough approximation.
The differences between the dynamics of simulated and real systems cause a ``reality gap'' that makes it unlikely for a simulation-trained policy to successfully transfer to a physical system \cite{Waheed_2025}.

In literature, researchers have put a significant effort into diminishing this gap \cite{Gilles_2025, Zhang_2025, Kim_2025}: to this aim, most of the approaches leverage domain randomization \cite{narang2022factory, andrychowicz2020learning, allshire2022transferring, rudin2022learning, chi_iterative_2022, chebotar_closing_2019} to expose the policy to a wide range of observation distributions in simulation, thus improving generalization onto a real system.
Nevertheless, naive domain randomization might not be sufficient to completely attenuate the dynamics gap: for instance, \cite{hwangbo_learning_2019} employs a specific network to mimic the real actuation system.
Another technique in this context is policy-level action integrator (PLAI) \cite{tang2023industreal}, a simple yet effective algorithm aimed at compensating the sim-to-real dynamic discrepancies with an integral action, which proved to be paramount for a successful transfer.

In this work, we employ sim-to-real strategies to deploy a policy on a real system; to the best of the authors' knowledge, this work represents the first instance of deploying PBRL agents on real hardware.

\subsection{Contribution}

This paper investigates a population-based reinforcement learning (PBRL) framework against robotics tasks, allowing to train a population of agents by exploiting GPU-based massively-parallel simulation in order to tune the hyperparameters during training.
We evaluate the PBRL framework on 4 complex tasks available in Isaac Gym \cite{makoviychuk2021isaac}: \textit{Anymal Terrain}, \textit{Humanoid}, \textit{Shadow Hand}, and \textit{Franka Nut Pick} (Fig.~\ref{fig:envs}).
Our results show that training a population of agents yields better performance compared to a single-agent baseline on all tasks.
The comparison is provided across 3 RL algorithms (PPO, SAC, and DDPG), varying the number of agents in a population, and across different hyperparameter mutation schemes. 
Finally, we deploy the PBRL agents on a real Franka Panda robot for a \textit{Franka Nut Pick} task, without any policy adaptation phase on the physical system.
In summary, the main contributions of this work are:

\begin{itemize} 
    \item a population-based RL framework that utilizes GPU-accelerated simulation to train robotic manipulation tasks by adaptively optimizing the set of hyperparameters during training;
    \item simulations demonstrating the effectiveness of the PBRL approach on 4 tasks using 3 RL algorithms, including both on-policy and off-policy methods, investigating the performance \wrt the number of agents and mutation mechanisms;
    \item sim-to-real transfer assessment of PBRL policies onto a real Franka Panda robot;
    \item an open-source codebase to train policies using the PBRL algorithm: \url{https://github.com/Asad-Shahid/PBRL}
\end{itemize}

\begin{comment}
\begin{figure*}[t]
    \centering
    \includegraphics[width=\textwidth]{Figures/approach.png}
    \captionsetup{width=\textwidth}
    \caption{PBRL framework used to learn robotic manipulation tasks through a combination of RL, evolutionary selection, and GPU-based parallel simulations.}
    \label{fig:approach}
\end{figure*}
\end{comment}

% Size of blocks (learning and simulation)
\newlength{\blockwidth}
\setlength{\blockwidth}{0.31\textwidth}
\newlength{\blockheight}
\setlength{\blockheight}{0.8\blockwidth}

% Height of inner block
\newlength{\innerheight}
\setlength{\innerheight}{0.75\blockheight}

% Horizontal offset: how much the left side of inner block is offset wrt the left side of outer block
\newlength{\ho}
\setlength{\ho}{0.025\blockwidth}

% Vertical offset: how much the bottom side of inner block is offset wrt the bottom side of outer block
\newlength{\vo}
\setlength{\vo}{0.05\blockheight}

% Width of the arrows between inner blocks
\newlength{\arrowwidth}
\setlength{\arrowwidth}{0.15\blockwidth}

% Width of inner blocks
\newlength{\innerwidth}
\setlength{\innerwidth}{\dimexpr 0.5\blockwidth - \ho - 0.5\arrowwidth \relax}

% Center of blocks (in absolute coordinates wrt to block center)
\newlength{\innerblockcx}
\setlength{\innerblockcx}{\dimexpr 0.5\arrowwidth + 0.5\innerwidth \relax}
\newlength{\innerblockcy}
\setlength{\innerblockcy}{\dimexpr 0.5\blockheight - \vo - 0.5\innerheight} \relax

% Bottom-left corner of left block (agents)
\newlength{\innerblockblx}
\setlength{\innerblockblx}{\dimexpr -\innerblockcx -0.5\innerwidth \relax}
\newlength{\innerblockbly}
\setlength{\innerblockbly}{\dimexpr -\innerblockcy -0.5\innerheight \relax}

% Text offset
\newlength{\textoffset}
\setlength{\textoffset}{\dimexpr 0.5\blockheight - 0.5\vo - 0.5\innerheight \relax}

% Image size
\newlength{\imagewidth}
\settowidth{\imagewidth}{\includegraphics{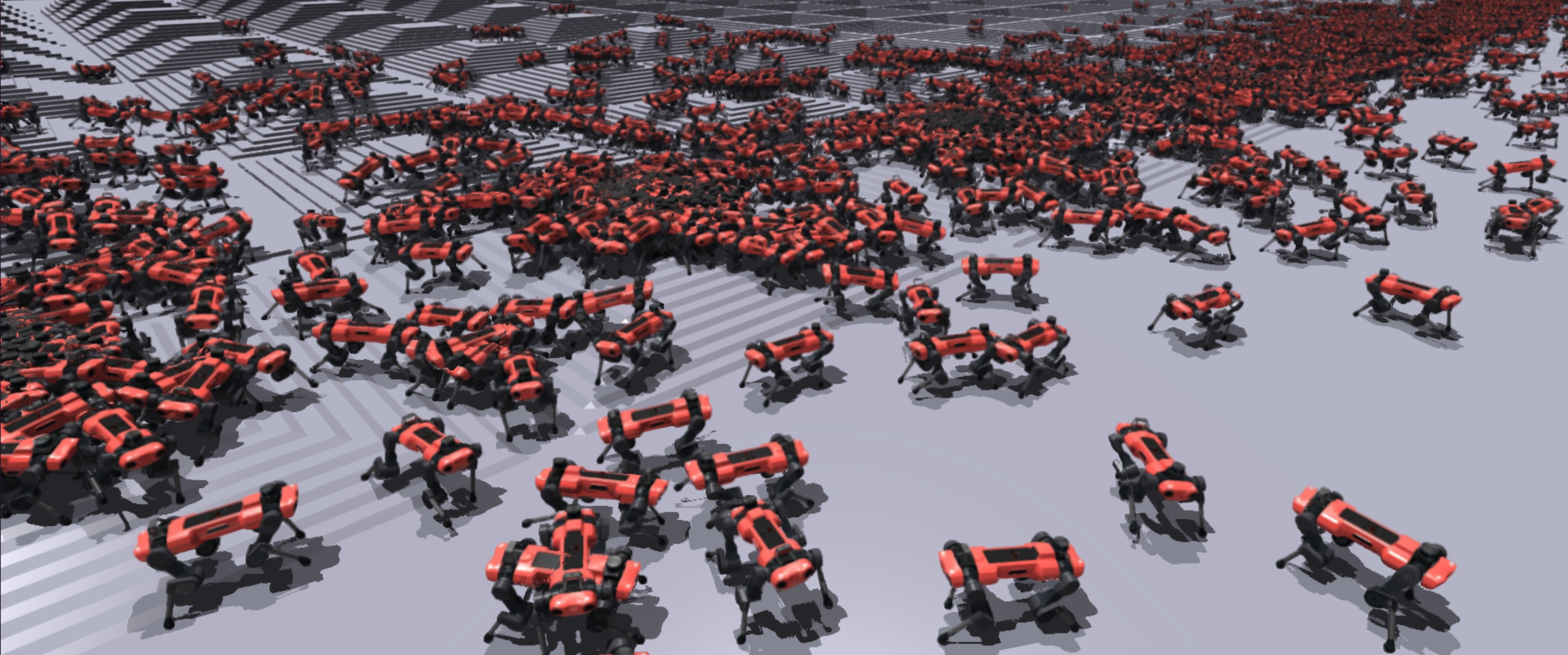}}
\newlength{\imageheight}
\settoheight{\imageheight}{\includegraphics{Figures/anymal.png}}

% How much the image must be trimmed to fit in the inner block
\newlength{\widthtotrim}
\setlength{\widthtotrim}{\innerwidth * \ratio{\imageheight}{\innerheight}}
\newlength{\lefttrim}
\setlength{\lefttrim}{260pt}
\newlength{\righttrim}
\setlength{\righttrim}{\dimexpr \imagewidth - \lefttrim - \widthtotrim \relax}

% Height of square inside Agents inner block
\newlength{\squareheight}
\setlength{\squareheight}{0.6\innerheight}

% Width of square inside Agents inner block
\newlength{\squarewidth}
\setlength{\squarewidth}{\dimexpr \innerwidth - 2\ho \relax}

% Center of square
\newlength{\squarecx}
\setlength{\squarecx}{-\innerblockcx}
\newlength{\squarecy}
\setlength{\squarecy}{\dimexpr -\innerblockcy -0.5\innerheight +\ho + 0.5\squareheight \relax}

% Bottom-left corner of square block (agents)
\newlength{\squareblx}
\setlength{\squareblx}{\dimexpr \squarecx -0.5\squarewidth \relax}
\newlength{\squarebly}
\setlength{\squarebly}{\dimexpr \squarecy -0.5\squareheight \relax}

% Policies text offset
\newlength{\ptextoffset}
\setlength{\ptextoffset}{\dimexpr 0.5\innerheight - 0.5\squareheight - 0.5\ho \relax}

% Ranking block size
\newlength{\rankingblockwidth}
\setlength{\rankingblockwidth}{0.65\blockwidth}
\newlength{\rankingblockheight}
\setlength{\rankingblockheight}{\innerheight}

% Transfer block size
\newlength{\transferblockwidth}
\setlength{\transferblockwidth}{0.2\blockwidth}
\newlength{\transferblockheight}
\setlength{\transferblockheight}{0.7\rankingblockheight}

% Mutate block size
\newlength{\mutateblockwidth}
\setlength{\mutateblockwidth}{\transferblockwidth}
\newlength{\mutateblockheight}
\setlength{\mutateblockheight}{0.25\transferblockheight}

\tikzstyle{block} = [draw, very thick, rounded corners=10pt, rectangle, minimum width=\blockwidth, minimum height=\blockheight]

\begin{figure*}[t]
\centering
\begin{tikzpicture}[auto,>=latex']
    % Learning block
    \node [block] (learning) {};
    \node [above, font=\normalsize] at (learning.north) {LEARNING};
    \node [font=\normalsize] at ([xshift=-\innerblockcx, yshift=-\textoffset] learning.north) {Agents};
    \node [font=\normalsize] at ([xshift=+\innerblockcx, yshift=-\textoffset] learning.north) {Simulation};
    \draw [black, thick] (learning) ++(\innerblockblx, \innerblockbly) rectangle ++(\innerwidth, \innerheight) node (agents) {};
    \node[inner sep=0pt] (simulation) at ([xshift=\innerblockcx, yshift=-\innerblockcy] learning.center) {\includegraphics[width=\innerwidth, height=\innerheight, trim=\lefttrim{} 0 \righttrim{} 0, clip]{Figures/anymal.png}};  %trim options: lbrt

    % Learning arrows
    \draw [->, thick] ([yshift=-1/3*\innerheight] agents.center) -- ([yshift=0.5\innerheight-1/3*\innerheight] simulation.west) node [midway, font=\normalsize] {$a_t$};
    \draw [->, thick] ([yshift=-0.5\innerheight+1/3*\innerheight] simulation.west) -- ([yshift=-2/3*\innerheight] agents.center) node [midway, font=\normalsize, align=center] {$s_t, r_t$};

    % Policies square
    \draw [black] (learning) ++(\squareblx, \squarebly) rectangle ++(\squarewidth, \squareheight) node[xshift=-0.5\squarewidth, yshift=-0.5\squareheight, align=center, font=\normalsize] (policies) {$\pi(\theta_1, h_1)$\\$\vdots$\\$\pi(\theta_n, h_n)$};
    \node [align=center, font=\normalsize] at ([xshift=0, yshift=0.5\squareheight+\ptextoffset] policies) {PPO, SAC,\\DDPG};

    % Evolution block
    \node [block, right=0.5\blockwidth of learning] (evolution) {};
    \node [above, font=\normalsize] at (evolution.north) {EVOLUTION};
    \node [draw, rounded corners=10pt, rectangle, thick, minimum width=\rankingblockwidth, minimum height=\rankingblockheight] (ranking) at ([xshift=2\ho+0.5\rankingblockwidth, yshift=\innerblockbly+0.5\rankingblockheight] evolution.west) {};
    \node [draw, rounded corners=10pt, rectangle, thick, minimum width=\transferblockwidth, minimum height=\transferblockheight] (transfer) at ([xshift=-2\ho-0.5\transferblockwidth, yshift=\innerblockbly+\innerheight-0.5\transferblockheight] evolution.east) {};
    \node [draw, rounded rectangle, font=\normalsize, minimum height=\mutateblockheight] (mutate) at ([xshift=0.5\blockwidth-2\ho-0.5\transferblockwidth, yshift=-0.5\blockheight+\vo+0.5\mutateblockheight] evolution) {Mutate};
    \node [font=\normalsize] at ([xshift=-0.5\blockwidth+2\ho+0.5\rankingblockwidth, yshift=-\textoffset] evolution.north) {Ranking};
    \node [font=\normalsize] at ([xshift=0.5\blockwidth-2\ho-0.5\transferblockwidth, yshift=-\textoffset] evolution.north) {Transfer};

    % Ranking block
    \node [draw, rectangle, rounded corners=10pt, minimum width=0.8\rankingblockwidth, minimum height=0.25\rankingblockheight, font=\normalsize] (bottomagents) at ([yshift=0.3\rankingblockheight] ranking) {Agents bottom 25\%};
    \node [draw, rectangle, rounded corners=10pt, minimum width=0.8\rankingblockwidth, minimum height=0.25\rankingblockheight, font=\normalsize] (midagents) at (ranking) {Agents mid 50\%};
    \node [draw, rectangle, rounded corners=10pt, minimum width=0.8\rankingblockwidth, minimum height=0.25\rankingblockheight, font=\normalsize] (topagents) at ([yshift=-0.3\rankingblockheight] ranking) {Agents top 25\%};

    % Transfer block
    \node [draw, rectangle, minimum width=0.5\transferblockwidth, font=\normalsize] (hyperparameters) at ([yshift=0.5\transferblockheight-0.5\rankingblockheight] transfer) {$h^*$};
    \node [font=\normalsize] (parameters) at ([yshift=-0.5\transferblockheight+0.5\rankingblockheight] transfer) {$\theta^*$};
            
    % Arrows
    \draw[->, very thick] ([yshift=0.5\blockheight-\vo-0.2\rankingblockheight] learning.east) -- node[font=\normalsize, anchor=north] {$\mathcal R_{\text{mean}}$} ([yshift=0.5\blockheight-\vo-0.2\rankingblockheight] evolution.west);
    %\draw [->, very thick] (evolution) -- node[name=back] {} (learning);
    \draw [->, very thick] (topagents.west) -- node[align=center, pos=0.6, font=\normalsize, anchor=south] (return) {Continue\\Learning} ([yshift=-0.5\blockheight+\vo+0.2\rankingblockheight] learning.east);
    \node [right of=topagents, node distance=0.45\rankingblockwidth, coordinate, font=\scriptsize] (topagentseast) {};
    \draw [-, thick] (topagents.east) -- (topagentseast);
    \draw [->, thick] (topagentseast) |- ([yshift=-0.5\transferblockheight+0.2\rankingblockheight] transfer.west);
    \node [left of=topagents, node distance=0.45\rankingblockwidth, coordinate, font=\scriptsize] (topagentswest) {};
    \draw [-, very thick] (midagents.west) -| (topagentswest);
    \draw [->, thick] ([yshift=0.5\transferblockheight-0.2\rankingblockheight] transfer.west) -- (bottomagents.east);
    \draw [->, thick] (mutate.north) -- (hyperparameters.south);
\end{tikzpicture}
\caption{The PBRL framework learns robotic tasks through a combination of RL, evolutionary selection, and GPU-based parallel simulations.}
\label{fig:approach}
\end{figure*}

\section{Methods}

This section describes the core concepts involved in the PBRL framework.
The overall approach, illustrated in Fig.~\ref{fig:approach}, can be viewed as a multi-layered training process consisting of an inner optimization loop with RL and an outer loop of online evolutionary selection with population-based training.
During training, the parameters of the agent's policy are updated at a higher rate using RL than the hyperparameters defining the RL procedure.

\subsection{Reinforcement Learning}\label{sec:rl}

The RL problem is modeled as a Markov Decision Process (MDP), where an agent interacts with the environment in order to maximize the expected sum of episodic rewards.
Specifically, a MDP is defined as $(\mathcal S, \mathcal A, \mathcal T, \mathcal  R, \gamma)$, where $\mathcal{S}$ is the set of states, $\mathcal{A}$ is the set of actions, $\mathcal T:\mathcal S \times \mathcal A \rightarrow \mathbb P(\mathcal S)$ is the transition dynamics, where $\mathbb P(\mathcal S)$ defines a probability distribution over $\mathcal S$, $\mathcal R:\mathcal S \times \mathcal A \rightarrow \mathbb R$ is the reward function, and $\gamma \in [0, 1]$ represents the discount factor.
The goal is formulated as learning a policy, either deterministic, $\pi_\theta: \mathcal{S} \rightarrow \mathcal{A}$, or stochastic, $\pi_\theta: \mathcal{S} \rightarrow \mathcal{D_A}$, where $\mathcal{D_A}$ represents a probability distribution over $\mathcal A$, and $\theta$ encapsulates the policy parameters, whose cardinality depends on the selected algorithm and network architecture.
In this work, the policy is learned using the on-policy method PPO, or either of the off-policy methods SAC or DDPG.
All these algorithms use an actor-critic architecture simultaneously learning the policy (actor) and the value function approximators (critics) $\mathcal{Q}:\mathcal{S \times A} \rightarrow \mathbb{R}$.
The implementation of critics in SAC and DDPG relies on double Q-learning and $n$-step returns.

To train the policy with PPO, a learning rate (LR) adaptation procedure is used based on a Kullback–Leibler (KL) divergence starting from an initial value $\eta_0$ \cite{makoviychuk2021isaac}.
At the end of each update iteration, the LR is increased by a factor of $K_\eta$ when the KL divergence between the current policy and the old policy is below the specified threshold, or reduced by $K_\eta$ if the KL divergence exceeds the threshold.

\begin{algorithm}
\caption{PBRL algorithm}\label{algo:pbrl}
\begin{algorithmic}[1]
\Require Initial population $\mathcal{P}$ of agents ($\Theta$ random, $H$ sampled from a uniform distribution)

\State $N_{\text{iter}} \gets 0$
\While{not end of training}
\State $\Theta \gets \text{Train} \big( \Pi(\Theta, H) \big)$ \Comment{Train all agents in $\mathcal{P}$}
\State $N_{\text{iter}} \gets N_{\text{iter}} + 1$
\If{$N_{\text{iter}} > N_{\text{start}}$ and $ N_{\text{iter}}$ \% $N_{\text{evo}} = 0$} 
\For{each agent $\pi(\theta_i, h_i) \in \mathcal P$}
\State $\mathcal R_{\text{mean}} \gets \text{Eval} \big( \pi(\theta_i, h_i) \big)$
\State Sort $\pi(\theta_i, h_i)$ based on $\mathcal R_{\text{mean}}$
\EndFor
\State Partition $\mathcal P$ into $\mathcal P_{\text{top}\,25\%}$, $\mathcal P_{\text{mid}\,50\%}$, $\mathcal P_{\text{bottom}\,25\%}$ 
\State $\mathcal P_{\text{bottom}\,25\%} \gets \varnothing$ 
\For{each agent $\pi(\theta_i^*, h_i^*) \in \mathcal P_{\text{top}\,25\%}$}
\State $h_i \gets \text{Mutate} (h_i^*)$ 
\State $\mathcal P_{\text{bottom}\,25\%} \gets \mathcal P_{\text{bottom}\,25\%} \cup \pi(\theta_i^*, h_i)$
\label{step:mutate}
\EndFor
\EndIf
\EndWhile
\end{algorithmic}
\end{algorithm}

In DDPG, the common practice involves adding a small noise to the deterministic actions of the policy to enable exploration. In this work, the noise is added following a mixed exploration strategy \cite{pmlr-v202-li23f}, where the general idea is akin to adding a different noise level for each environment when training in a massively-parallel regime.
For the $i$-th environment out of $N \in \mathbb N$ environments, a zero mean and uncorrelated Gaussian noise is given as: $\mathcal N(0, \sigma_i)$, where $\sigma_i = \sigma_{\text{min}} + \frac{i-1}{N-1} (\sigma_{\text{max}} - \sigma_{\text{min}})$, with $\sigma_{\text{min}}$ and $\sigma_{\text{max}}$ being hyperparameters.

\begin{table*}
\centering
\caption{Hyperparameters setup for PPO and PBRL-PPO across all the tasks.}
\label{tab:hp_ppo}
\begin{tabular}{c|ccc|ccc}
\multirow{2}{*}{\textbf{Hyperparameter}} & \multicolumn{3}{c|}{\textbf{PPO}} & \multicolumn{3}{c}{\textbf{PBRL-PPO}} \\
 & \textit{Anymal Terrain} & \begin{tabular}[c]{@{}c@{}}\textit{Shadow Hand \&}\\\textit{Humanoid}\end{tabular} & \textit{Franka Nut Pick} & \textit{Anymal Terrain} & \begin{tabular}[c]{@{}c@{}}\textit{Shadow Hand \&}\\\textit{Humanoid}\end{tabular} & \textit{Franka Nut Pick} \\ 
\hline
Environments per agent & 4096 & 16384 & 128 & 1024 & 4096 & 128 \\
MLP hidden units & [512, 256, 128] & [512, 256, 128] & [256, 128, 64] & [512, 256, 128] & [512, 256, 128] & [256, 128, 64] \\
Horizon & 32 & 16 & 120 & 32 & 16 & 120 \\
Batch size & 8192 & 32768 & 512 & 8192 & 8192 & 512 \\
Actor variance & 0.5 & 1 & 1 & 0.3 -- 1 & 0.3 -- 1 & 0.3 -- 1 \\
KL threshold & 0.016 & 0.016 & 0.016 & 0.008 -- 0.016 & 0.008 -- 0.016 & 0.008 -- 0.016 \\
Entropy loss coefficient & 0.001 & 0.001 & 0 & 0.0001 -- 0.001 & 0.0001 -- 0.001 & 0.0001 -- 0.001 \\
Epochs & 8 & 4 & 8 & 8 & 4 & 8 \\
Discount factor $\gamma$ & 0.99 & 0.99 & 0.99 & 0.99 & 0.99 & 0.99 \\
GAE lambda & 0.95 & 0.95 & 0.95 & 0.95 & 0.95 & 0.95 \\
PPO clip $\epsilon$ & 0.2 & 0.2 & 0.2 & 0.2 & 0.2 & 0.2 \\
Initial LR $\eta_0$ & \num{5e-4} & \num{5e-4} & \num{5e-4} & \num{5e-4} & \num{5e-4} & \num{5e-4} \\
LR adaptation gain $K_\eta$ & 1.5 & 1.5 & 1.5 & 1.5 & 1.5 & 1.5
\end{tabular}
\vspace*{-1em}
\end{table*}

\subsection{Population-Based Training}\label{sec:pbt}

While a standard RL agent aims to learn an optimal policy by interacting with an environment and iteratively updating the policy through an optimization method, PBRL uses a population $\mathcal P$ composed of $n \triangleq \lvert \mathcal P \rvert$ agents, each interacting with the environment independently to collect experience and learn its own policy\footnote{This setup differs from multi-agent reinforcement learning, where agents interact in a shared environment with a joint or competitive objective}.
Using evolutionary selection, the population is evaluated periodically based on a fitness metric, and best-performing members replace the worst-performing members, \ie, weights of the best agents are copied over, along with the mutated hyperparameters.
While this process does not offer a theoretical guarantee of reaching the global optimum --- a cumbersome objective to achieve through naive trial-and-error tuning --- evolutionary algorithms have demonstrated a strong empirical tendency to converge to the optimum across a wide range of domains \cite{jaderberg2017population,Lehman_2011,parker2020provably}.

In this work, a specific PBRL approach, population-based training (PBT), is employed as an outer optimization loop to enable diverse exploration and dynamically adapt the hyperparameters in high-dimensional continuous control tasks. 
The $i$-th agent $\pi(\theta_i, h_i) \in \mathcal P$ is characterized by the sets $\theta_i$ and $h_i$, where $\theta_i$ contains the parameters of the policy, and $h_i$ contains the hyperparameters that are optimized during training.
To represent the whole population $\mathcal P$, we denote with $\Theta \triangleq \bigcup_{i=1}^n \theta_i$, $H \triangleq \bigcup_{i=1}^n h_i$ and $\Pi \triangleq \{\pi(\theta_i,h_i)\}_{i=1}^n$ the sets of all the parameters, hyperparameters and policies respectively.

\begin{table}
\centering
\caption{Hyperparameters setup for off-policy algorithms on all four tasks. \textsuperscript*For \textit{Franka Nut Pick} these parameters are, respectively: 128, [256, 128, 64], 512.}
\label{tab:hp_off}
\begin{tabular}{c|c|c}
\textbf{Hyperparameter} & \begin{tabular}[c]{@{}c@{}}\textbf{SAC \&}\\\textbf{DDPG}\end{tabular} & \begin{tabular}[c]{@{}c@{}}\textbf{PBRL-SAC \&}\\\textbf{PBRL-DDPG}\end{tabular} \\ 
\hline
Environments per agent\textsuperscript* & 2048 & 2048 \\
MLP hidden units\textsuperscript* & [512, 256, 128] & [512, 256, 128] \\
Batch size\textsuperscript* & 4096 & 4096 \\
Horizon & 1 & 1 \\
Target update rate $\tau$ & \num{5e-2} & \num{5e-2} \\
Actor learning rate & 0.0001 & 0.0001 -- 0.001 \\
Critic learning rate & 0.0001 & 0.0001 -- 0.001 \\
DDPG exploration $\sigma_{\text{min}}$ & 0.01 & 0.01 -- 0.1 \\
DDPG exploration $\sigma_{\text{max}}$ & 1 & 0.5 -- 1 \\
SAC target entropy & -20 & -20 -- -10 \\
Replay buffer size & \num{1e6} & \num{1e6} \\
Epochs & 4 & 4 \\
$n$-step returns & 3 & 3
\end{tabular}
\vspace*{-1em}
\end{table}

Algorithm~\ref{algo:pbrl} provides pseudocode for the PBRL.
Training proceeds iteratively, where all agents are first independently trained by performing updates to the vector $\theta_i$.
After a certain number of policy updates $N_{\text{evo}}$ (each agent having been trained for some steps), the agents are evaluated and sorted based on the average return $\mathcal R_{\text{mean}}$ obtained over all of the previous episodes.
The agents in $\mathcal P_{\text{bottom}\,25\%}$ get replaced by randomly-sampled agents in $\mathcal P_{\text{top}\,25\%}$ with mutated hyperparameters, while the rest of the agents in $\mathcal P_{\text{mid}\,50\%}$ and $\mathcal P_{\text{top}\,25\%}$ continue training.

We consider 3 mutation mechanisms to generate the mutated hyperparameters (see line~\ref{step:mutate} of Algorithm~\ref{algo:pbrl}):
\begin{enumerate*}[label=(\roman*)]
    \item random perturbation is applied to the hyperparameters of the parent agent(s) through perturbation factors in Table \ref{tab:hp_pbrl};
    \item new hyperparameters are sampled from a prior uniform distribution with bounds specified in Table~\ref{tab:hp_ppo} and ~\ref{tab:hp_off};
    \item according to the DexPBT mutation scheme \cite{petrenko2023dexpbt}, hyperparameters are multiplied or divided by a random number $\mu$ sampled from a uniform distribution, \ie, $\mu \sim \mathcal U(\mu_{\text{min}}, \mu_{\text{max}})$ with probability $\beta_{\text{mut}} \in [0, 1]$
\end{enumerate*}.
Sect.~\ref{sec:mutation-comparison} compares all 3 mutation schemes.
After beginning the training, evolution is enabled after $N_{\text{start}} \in \mathbb N$ steps as in \cite{jaderberg2018human} to allow for initial exploration and promote population diversity.

\section{Experiments}
\label{sec:exp}

%\subsection{Environments} % Inlcude figures

The PBRL framework is evaluated on some of the most challenging benchmark tasks available in Isaac Gym, including \textit{Anymal Terrain}, \textit{Shadow Hand}, \textit{Humanoid}, and \textit{Franka Nut Pick} (Fig.~\ref{fig:envs}).
The experiments are conducted on a workstation with a single NVIDIA RTX 4090 GPU and \SI{32}{\giga\byte} of RAM.
Parallelizing the data collection across the GPU, Isaac Gym's PhysX engine can simulate thousands of environments using the above hardware.

\subsection{Results}
\label{sec:res}

\begin{table}
\centering
\caption{Parameter setup for PBRL.}
\label{tab:hp_pbrl}
\begin{tabular}{c|cc}
\multirow{2}{*}{\textbf{Parameter}} & \multicolumn{2}{c}{\textbf{Value}} \\
 & \textit{Franka Nut Pick} & Others \\ 
\hline
Evolution start $N_{\text{start}}$ & \num{2e5} steps & \num{1e7} steps \\
Evolution frequency $N_{\text{evo}}$ & \num{1e5} steps & \num{2e6} steps \\
Perturbation factor (min.) & 0.8 & 0.8 \\
Perturbation factor (max.) & 1.2 & 1.2
\end{tabular}
\vspace*{-1em}
\end{table}

\begin{figure*}
    \centering

    % Legend
    \begin{center}
    \footnotesize{
    {\legred}~Standard RL \hspace{0.3em}
    {\leggreen}~PBRL 4~agents %\par
    {\legblue}~PBRL 8~agents \hspace{0.3em}
    {\legpurple}~PBRL 16~agents}
    \end{center}

    % Figure
    \begin{subfigure}{0.24\textwidth}
        \centering
        \includegraphics[width=\textwidth]{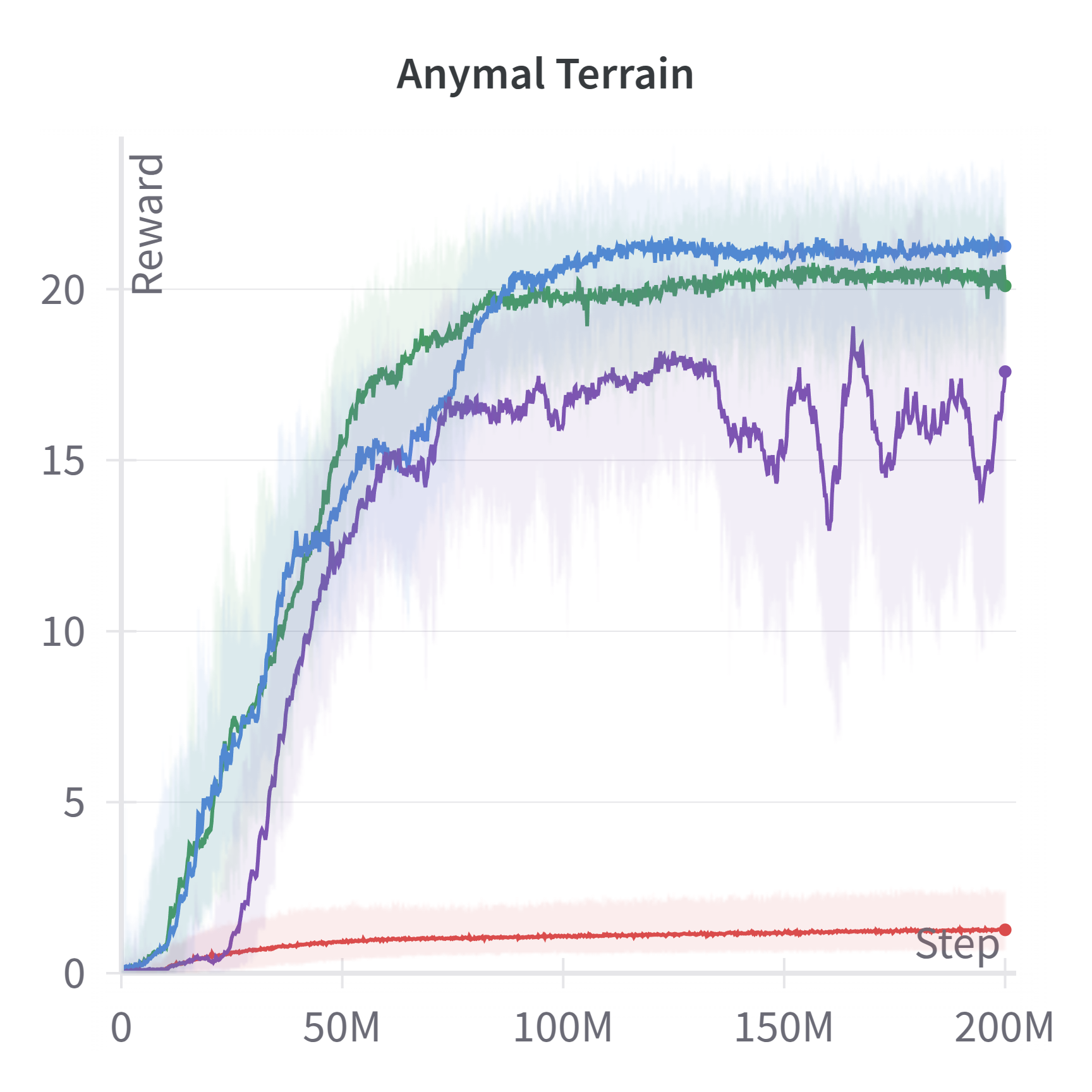}
    \end{subfigure}
    \hfill
    \begin{subfigure}{0.24\textwidth}
        \centering
        \includegraphics[width=\textwidth]{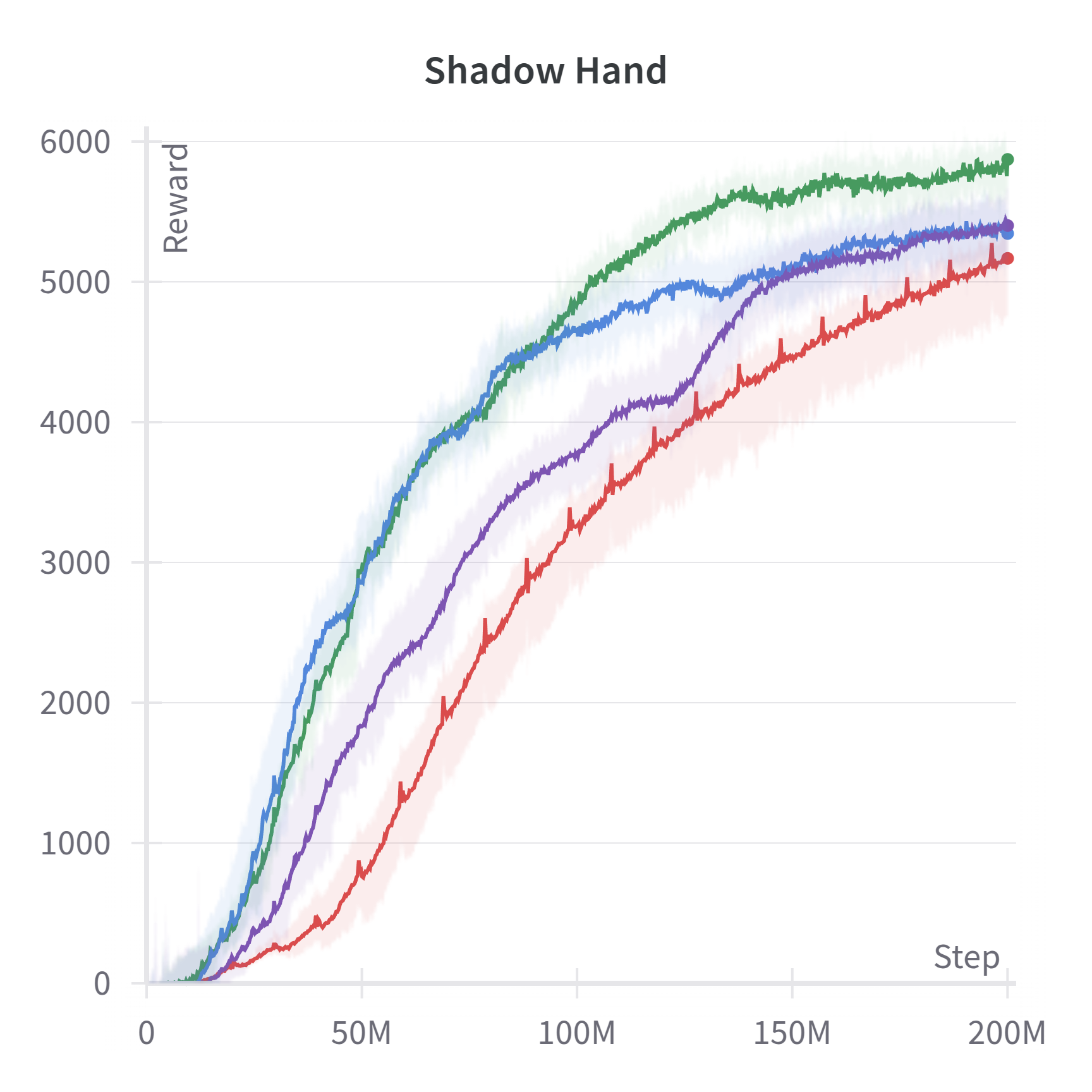}
    \end{subfigure}
    \hfill
    \begin{subfigure}{0.24\textwidth}
        \centering
        \includegraphics[width=\textwidth]{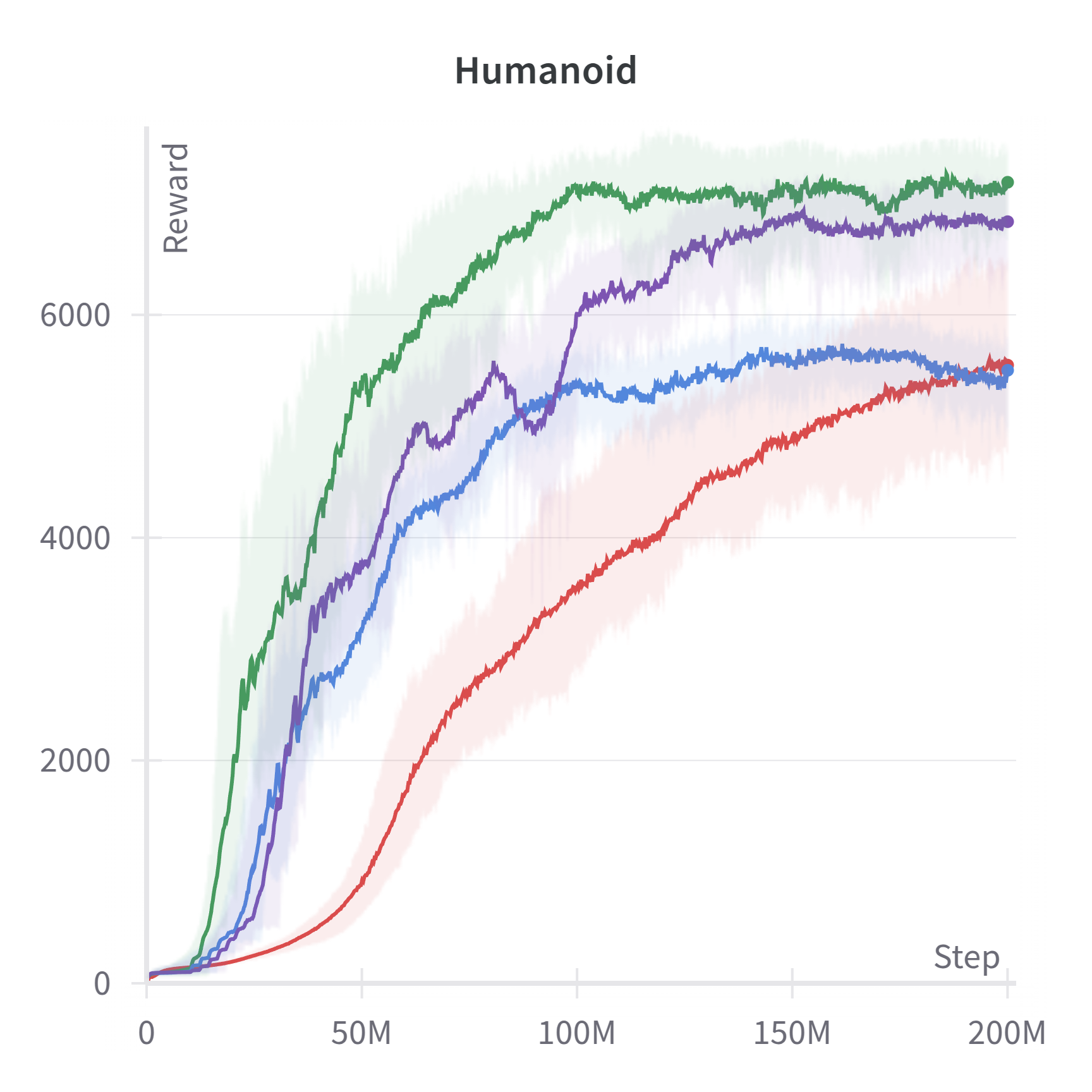}
    \end{subfigure}
    \hfill
    \begin{subfigure}{0.24\textwidth}
        \centering
        \includegraphics[width=\textwidth]{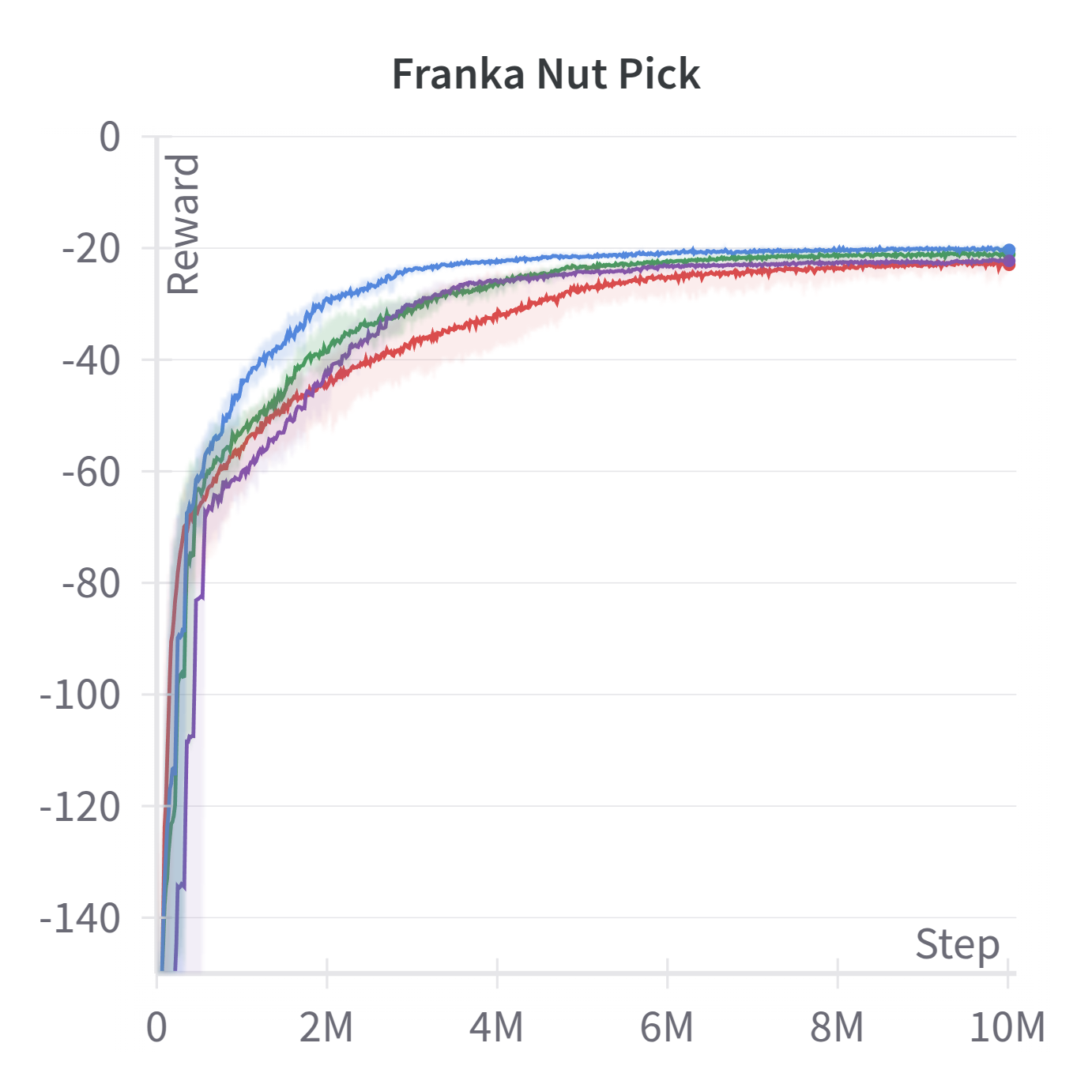}
    \end{subfigure}
    \\
    \begin{subfigure}{0.24\textwidth}
        \centering
        \includegraphics[width=\textwidth]{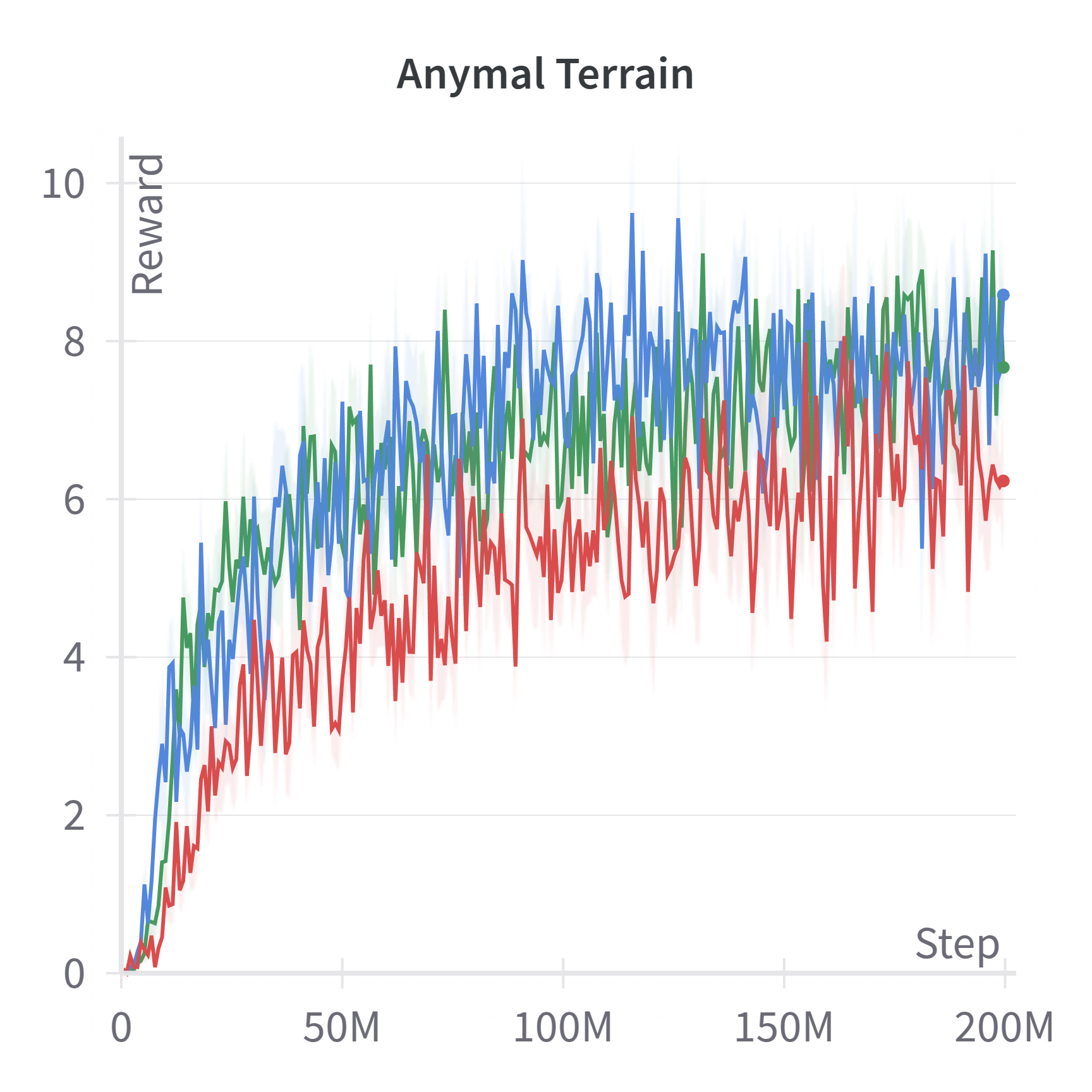}
    \end{subfigure}
    \hfill
    \begin{subfigure}{0.24\textwidth}
        \centering
        \includegraphics[width=\textwidth]{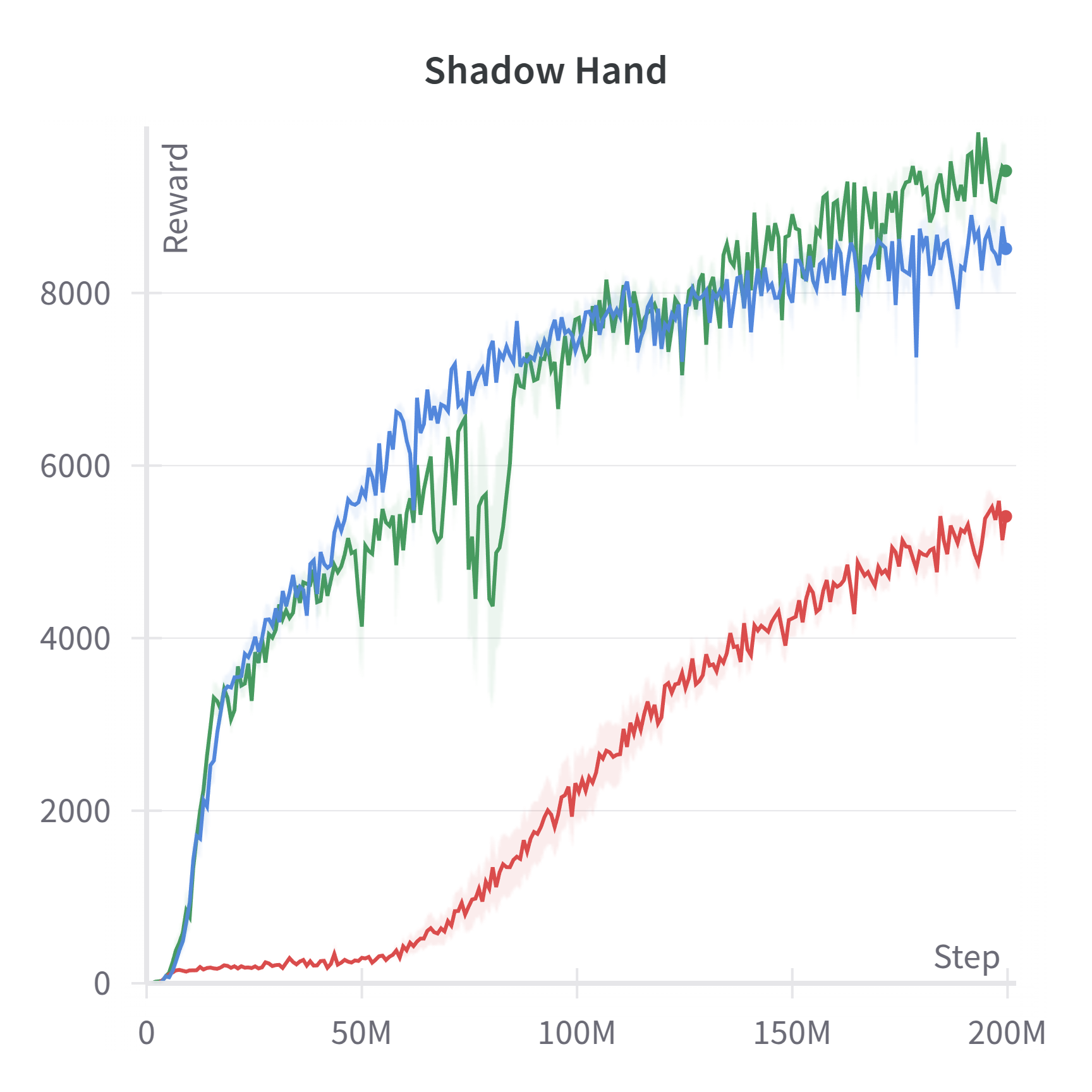}
    \end{subfigure}
    \hfill
    \begin{subfigure}{0.24\textwidth}
        \centering
        \includegraphics[width=\textwidth]{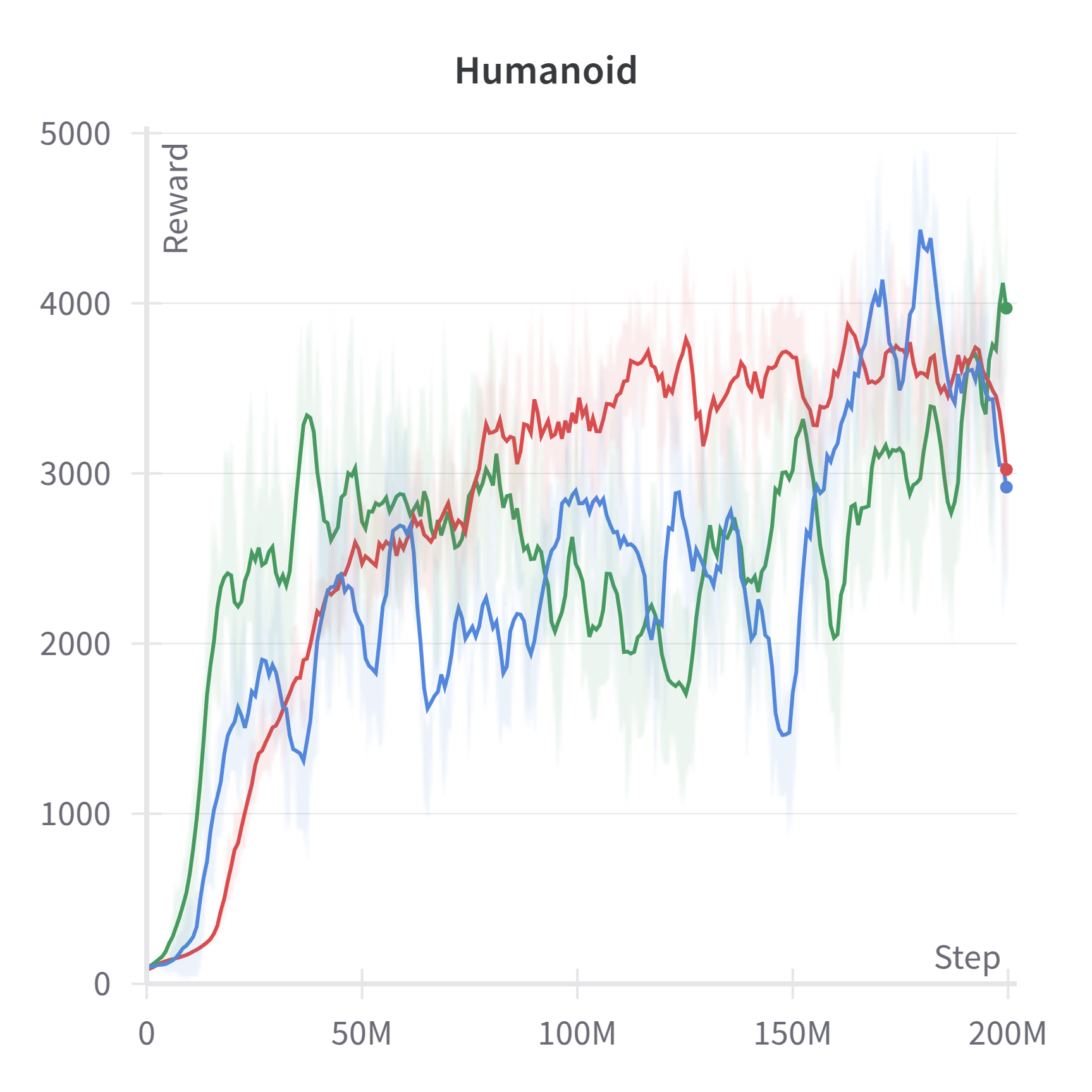}
    \end{subfigure}
    \hfill
    \begin{subfigure}{0.24\textwidth}
        \centering
        \includegraphics[width=\textwidth]{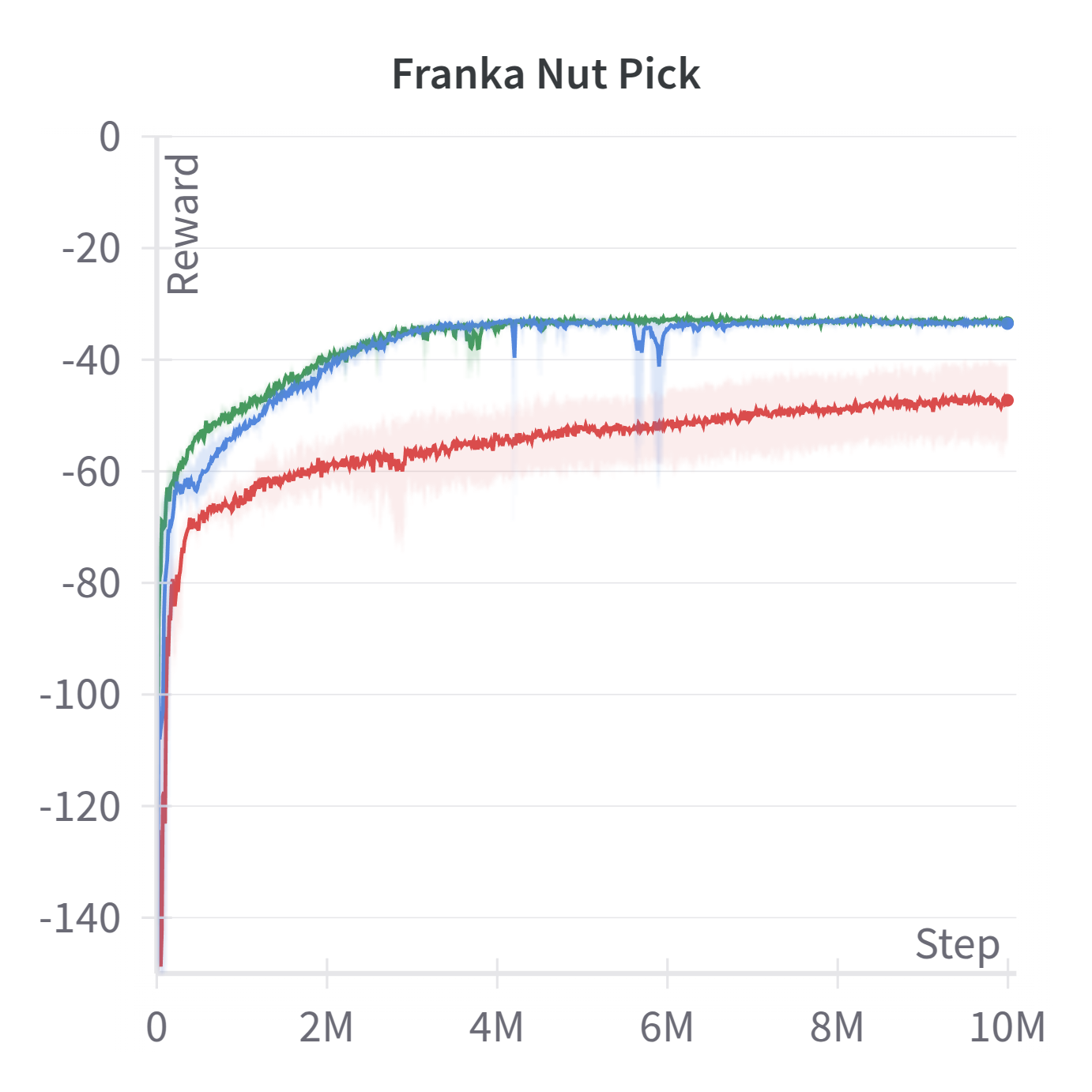}
    \end{subfigure}
    \\
    \begin{subfigure}{0.24\textwidth}
        \centering
        \includegraphics[width=\textwidth]{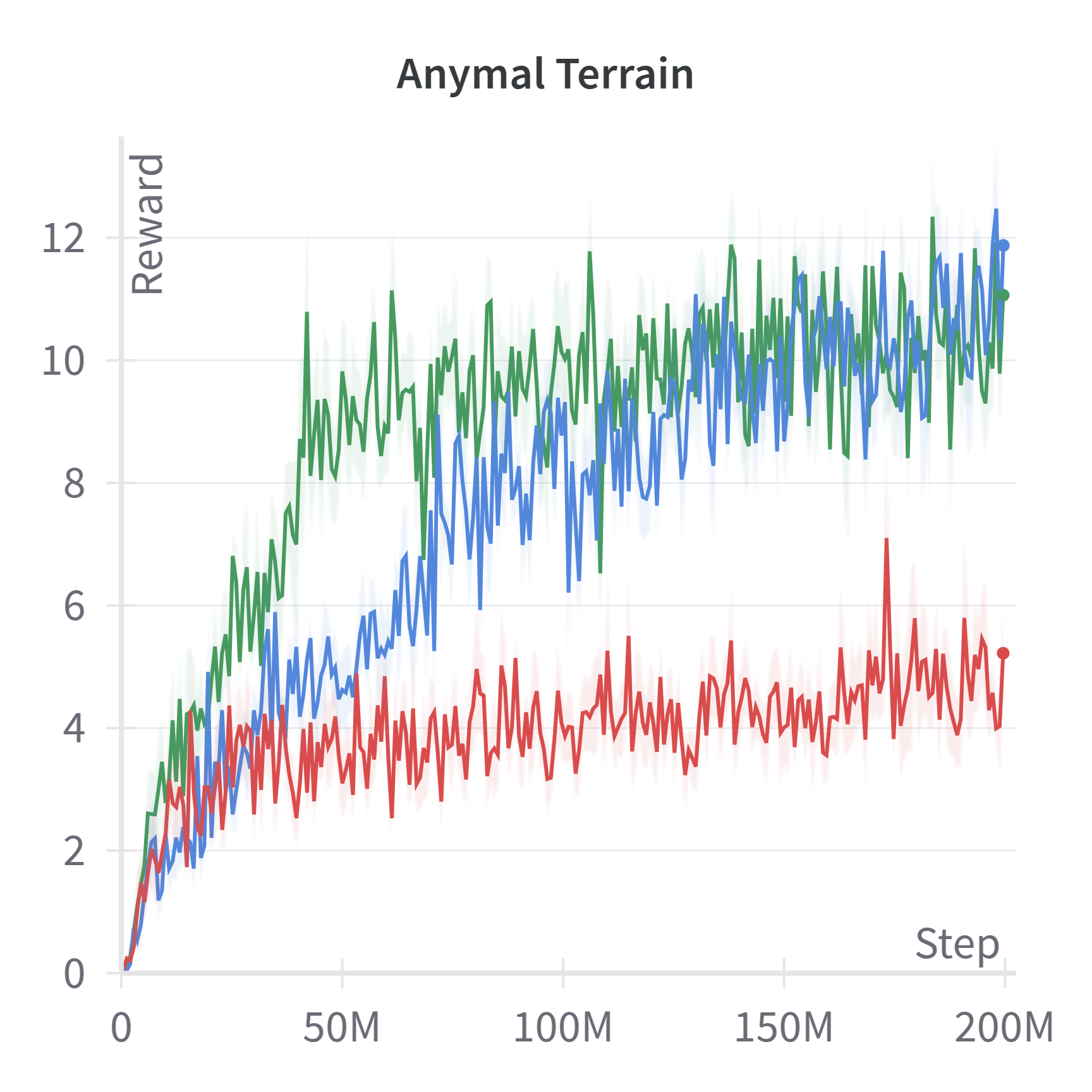}
    \end{subfigure}
    \hfill
    \begin{subfigure}{0.24\textwidth}
        \centering
        \includegraphics[width=\textwidth]{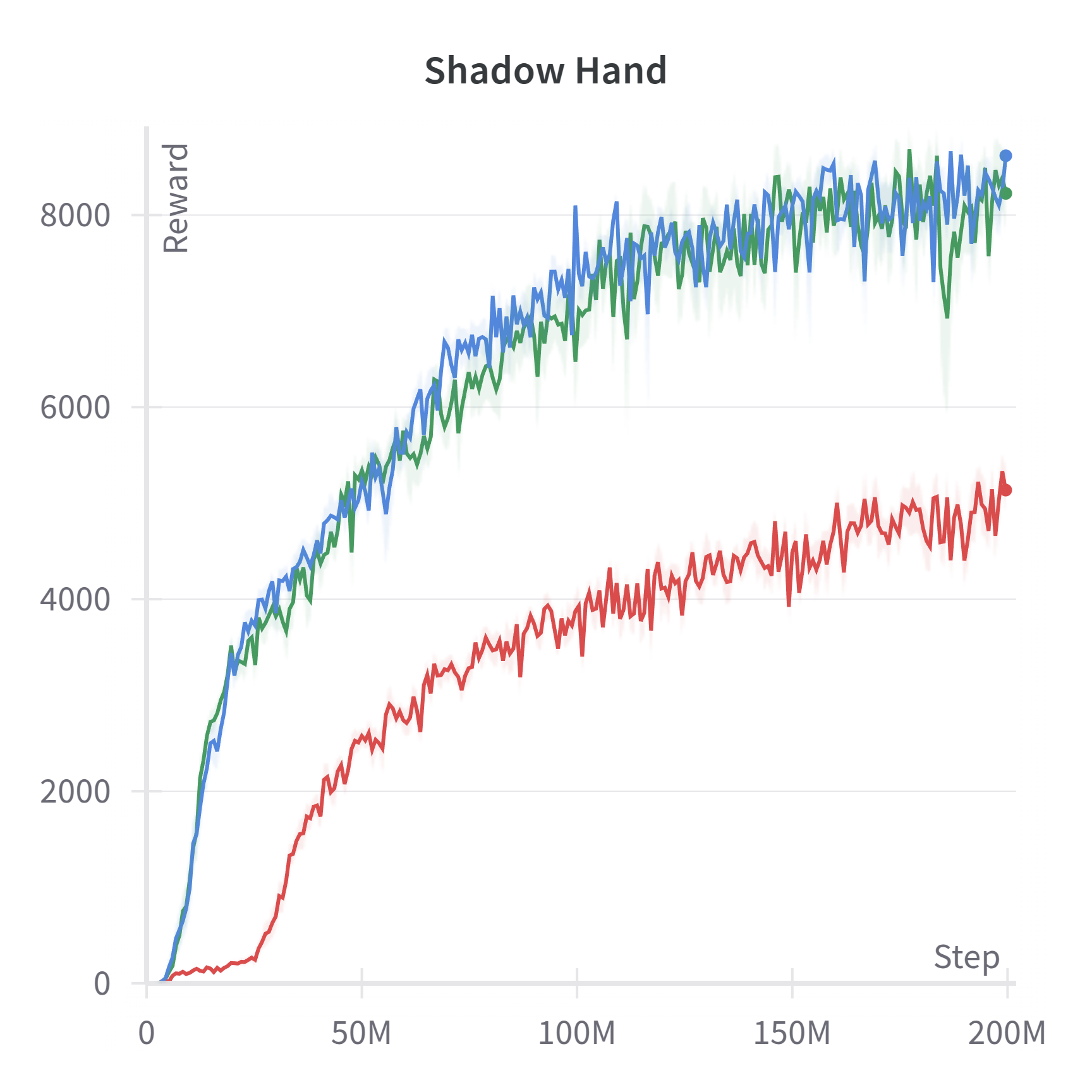}
    \end{subfigure}
    \hfill
    \begin{subfigure}{0.24\textwidth}
        \centering
        \includegraphics[width=\textwidth]{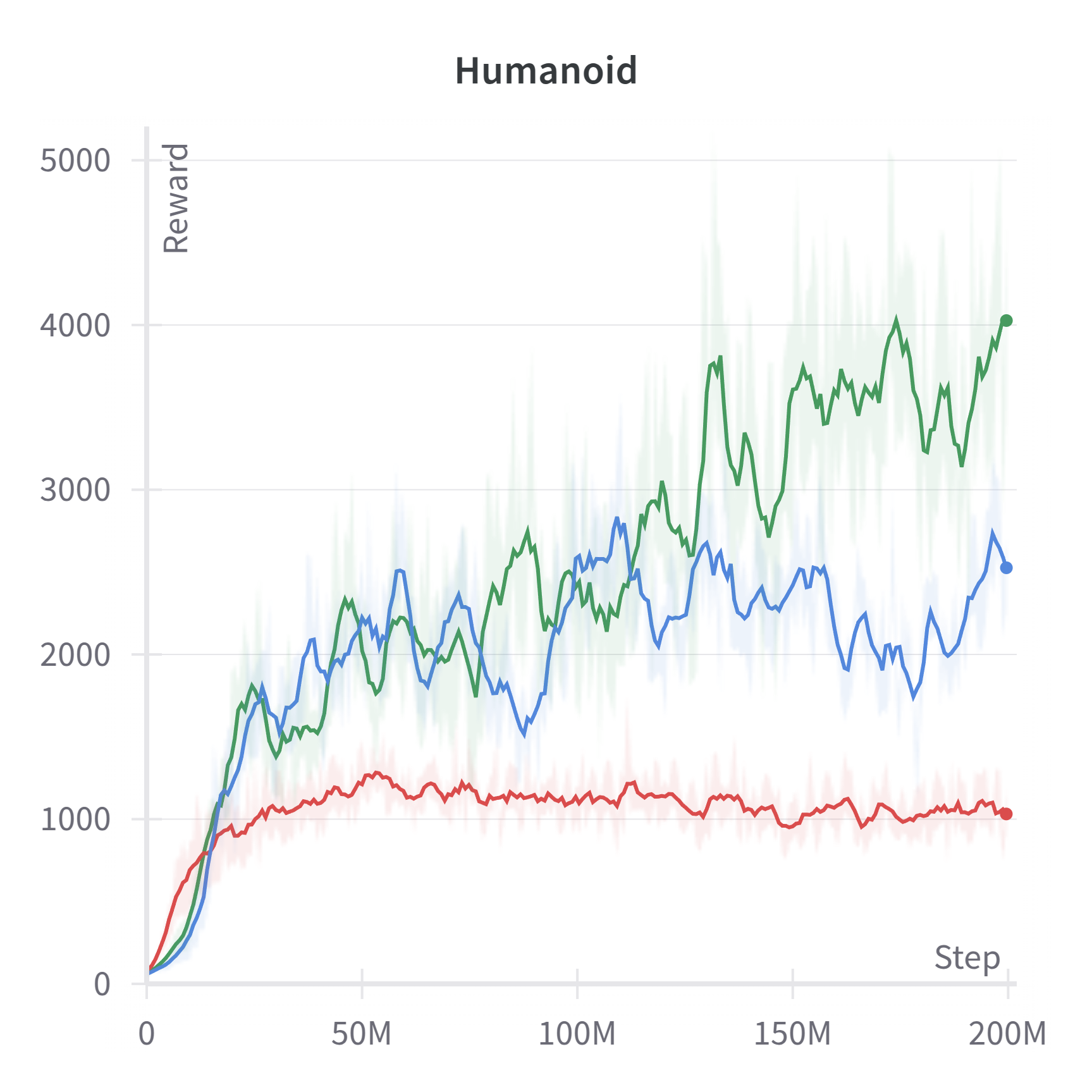}
    \end{subfigure}
    \hfill
    \begin{subfigure}{0.24\textwidth}
        \centering
        \includegraphics[width=\textwidth]{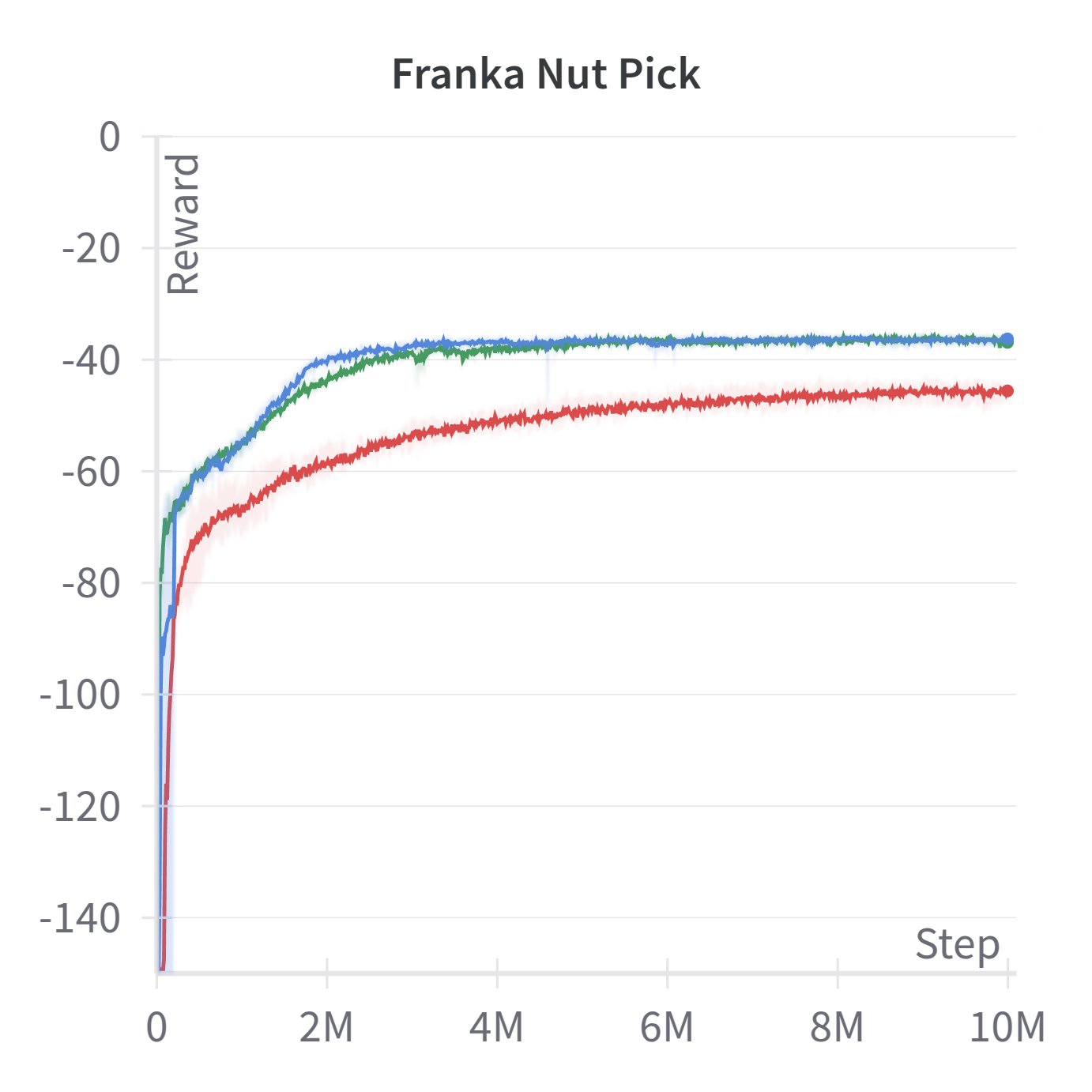}
    \end{subfigure}
    
    % Caption
    %\captionsetup{width=\columnwidth}
    \caption{
    Training results of baseline PPO (top), SAC (middle), and DDPG (bottom), along with their PBRL counterparts for $\lvert \mathcal P \rvert \in \{ 4, 8, 16 \}$.
    The shaded area shows the standard deviation around the mean performance across agents in $S$, or among 8 seeds in non-evolutionary baselines.
    SAC and DDPG are not evaluated on 16 agents due to higher memory usage.
    }
    \label{fig:pbrl}
    \vspace*{-1.5em}
\end{figure*}

The experiments focus on optimizing the hyperparameters of the RL agents in a population and comparing the results against non-evolutionary baseline agents.
For each case of baseline agents, 8 experiments are run with different seeds.
Table \ref{tab:hp_ppo} and \ref{tab:hp_off} provide the hyperparameters for on-policy and off-policy algorithms, listing the sampling ranges of those optimized through the PBRL Algorithm~\ref{algo:pbrl}.
For a fair evaluation, we evaluate PBRL on at least 8 agents in total, \ie, for $\lvert \mathcal P \rvert \in \{ 1, 4, 8, 16 \}$, $S \triangleq (8, 2, 1, 1)$ independent simulations are run, respectively\footnote{With a slight abuse of notation, $\lvert \mathcal P \rvert = 1$ indicates a base RL agent}.
Our benchmarking campaign consists in evaluating the agents' reward comparing the independent (standard RL) and the evolutionary (PBRL) scenarios, to assess whether agents achieve higher reward or faster convergence when trained in populations.

\subsubsection{PBRL-PPO}

For the PPO agents, the tuned hyperparameters are the KL divergence threshold for an adaptive LR, the entropy loss coefficient, and the variance of action selection.
These parameters are crucial in ensuring sufficient exploration of the environment.
Fig.~\ref{fig:pbrl} shows the learning curves for the single-agent PPO baseline and PBRL-PPO for $\lvert \mathcal P \rvert \in \{4, 8, 16\}$\footnote{Readers should be aware that one can only choose a \textit{single} agent in $\mathcal P$ at deployment time (\eg the best one), as done in Sect.~\ref{sec:simtoreal}: we compare the agents' \textit{mean} reward across $\mathcal P$ to ease the visual comparison}.
The results demonstrate that PBRL-PPO outperforms PPO on 3 out of 4 tasks, yielding a higher return, with significant improvement seen in \textit{Anymal Terrain}, which involves traversing increasingly challenging terrains.
For \textit{Franka Nut Pick}, PBRL agents match baseline PPO performance, as in this relatively straightforward task randomization alone suffices for a thorough exploration of the state/action space.

\subsubsection{PBRL-SAC}

The hyperparameters optimized in PBRL-SAC include the LR of the actor-critic networks and the target entropy factor to control the exploration behavior.
Entropy is a central theme in SAC agents as the policy is trained to maximize the trade-off between the expected return and the entropy.
The experiments in PBRL-SAC are run with a population size of $\lvert \mathcal P \rvert \in \{4, 8\}$.
Off-policy replay buffers require more memory allocation on GPU than on-policy methods.
Since all agents in PBRL-SAC train with their individual replay buffers, the maximum population size is set to 8 due to the limited GPU memory.
Fig.~\ref{fig:pbrl} displays SAC and PBRL-SAC training performance.
PBRL-SAC improves the training performance compared to non-evolutionary SAC on 3 out of 4 tasks, yielding a remarkable improvement on both \textit{Shadow Hand} and \textit{Franka Nut Pick}, while comparable results are achieved on \textit{Humanoid}, probably due to the limited task complexity.

\begin{figure*}
    \centering
    \begin{center}
    \footnotesize{
    {\legred}~Perturbation \hspace{0.3em}
    {\leggreen}~DexPBT \cite{petrenko2023dexpbt}} \hspace{0.3em}
    {\legblue}~Uniform Sampling
    \end{center}

    % Figure
    \begin{subfigure}{0.24\textwidth}
        \centering
        \includegraphics[width=\textwidth]{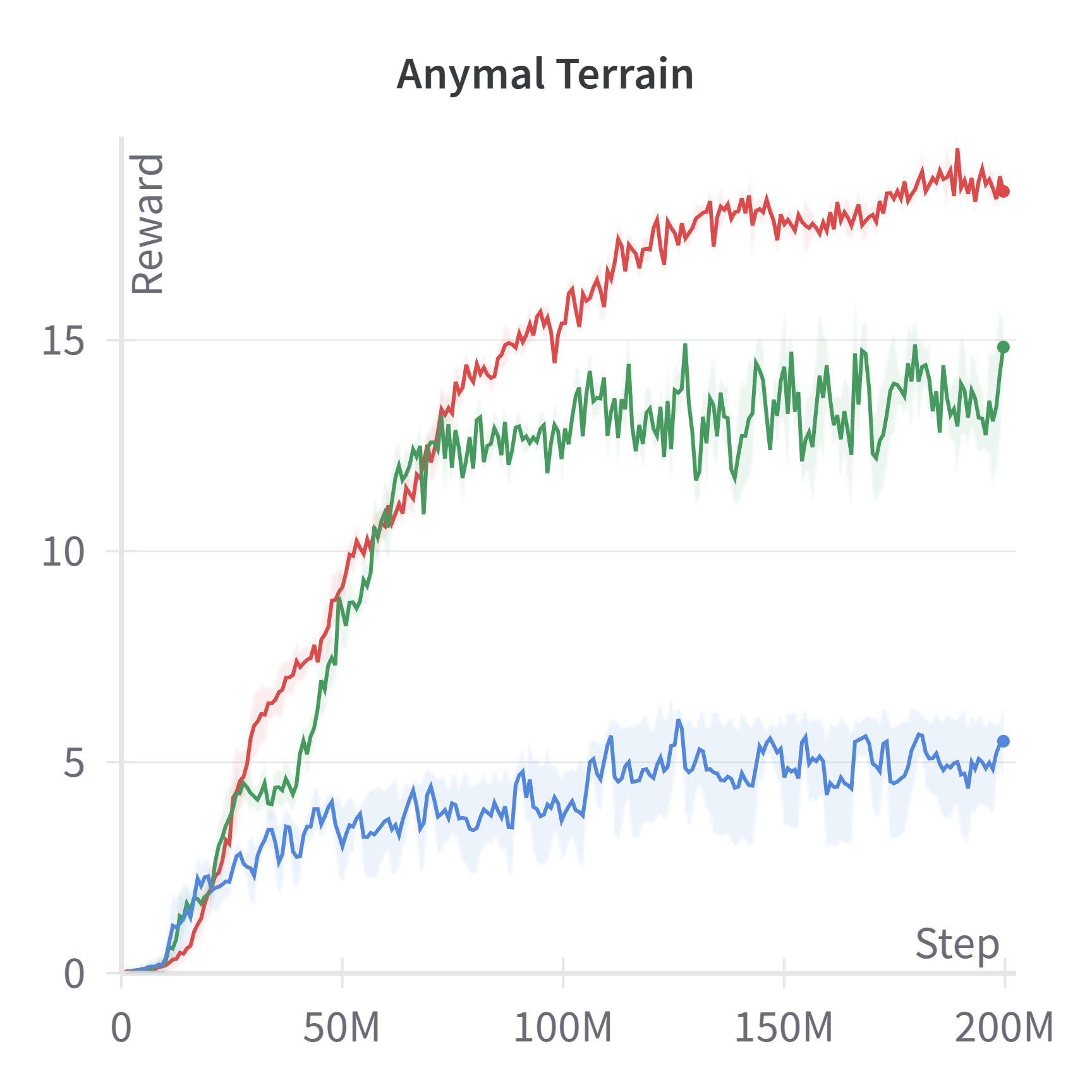}
    \end{subfigure}
    \hfill
        \begin{subfigure}{0.24\textwidth}
        \centering
    \includegraphics[width=\textwidth]{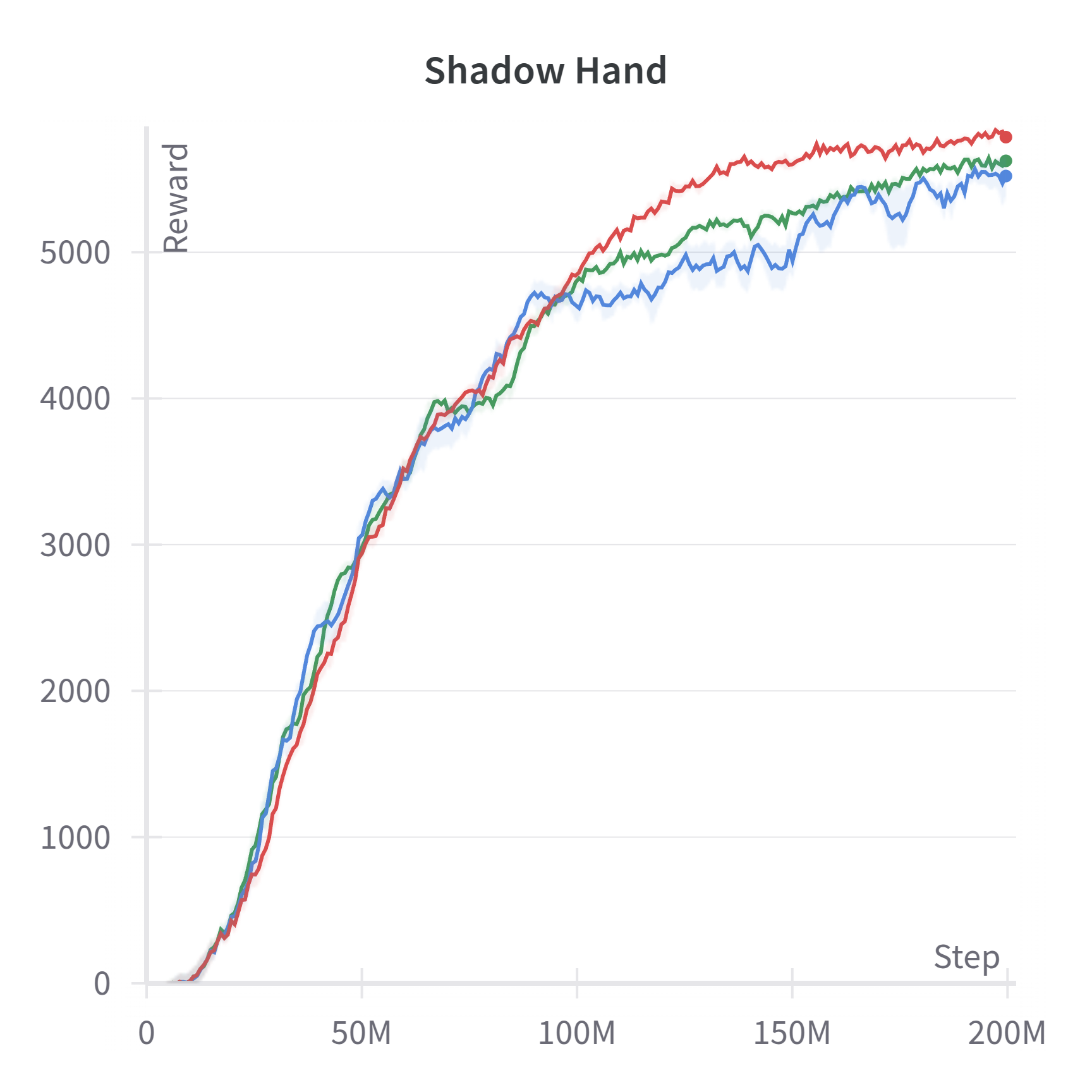}
    \end{subfigure}
    \hfill
    \begin{subfigure}{0.24\textwidth}
        \centering
        \includegraphics[width=\textwidth]{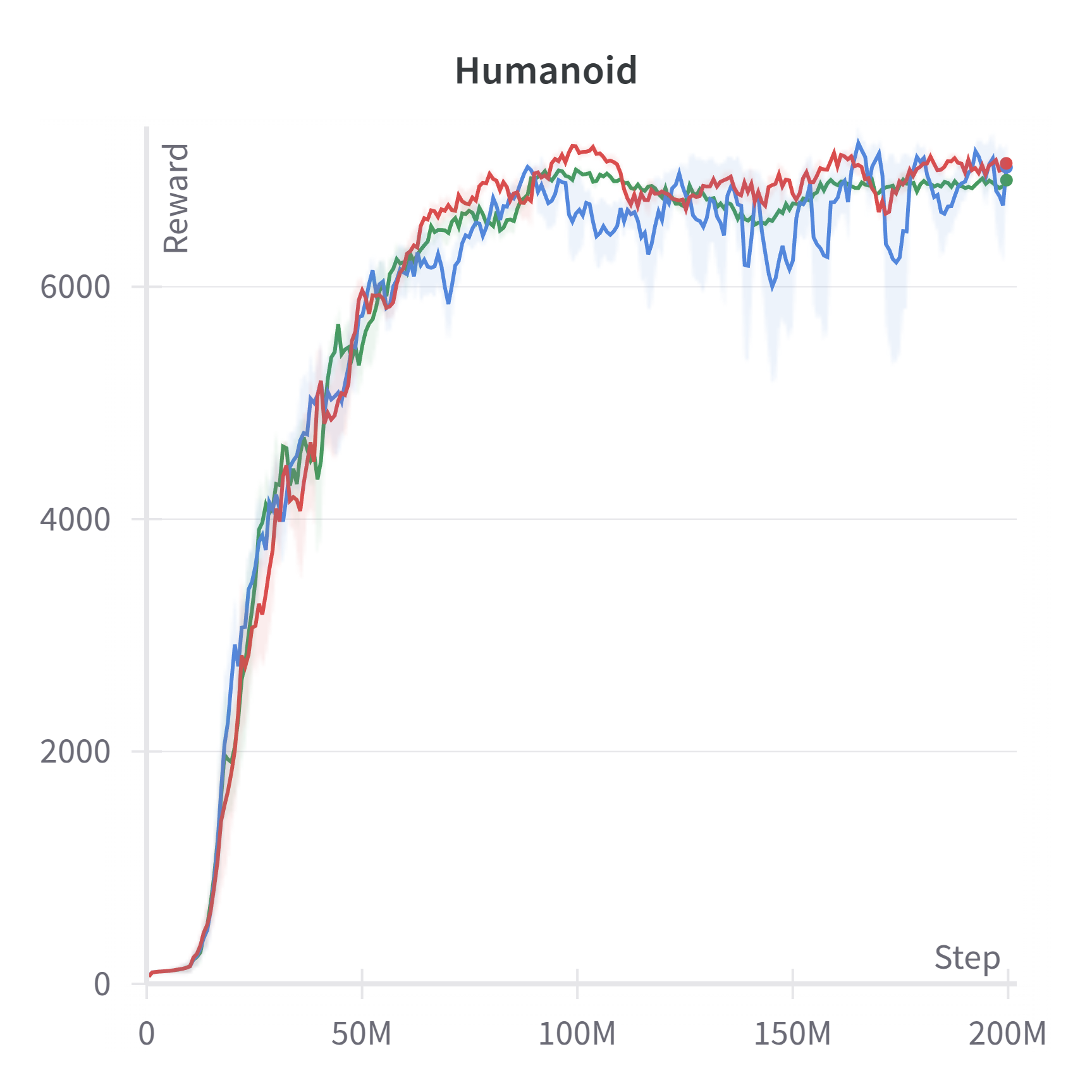}
    \end{subfigure}
    \hfill
        \begin{subfigure}{0.24\textwidth}
        \centering
    \includegraphics[width=\textwidth]{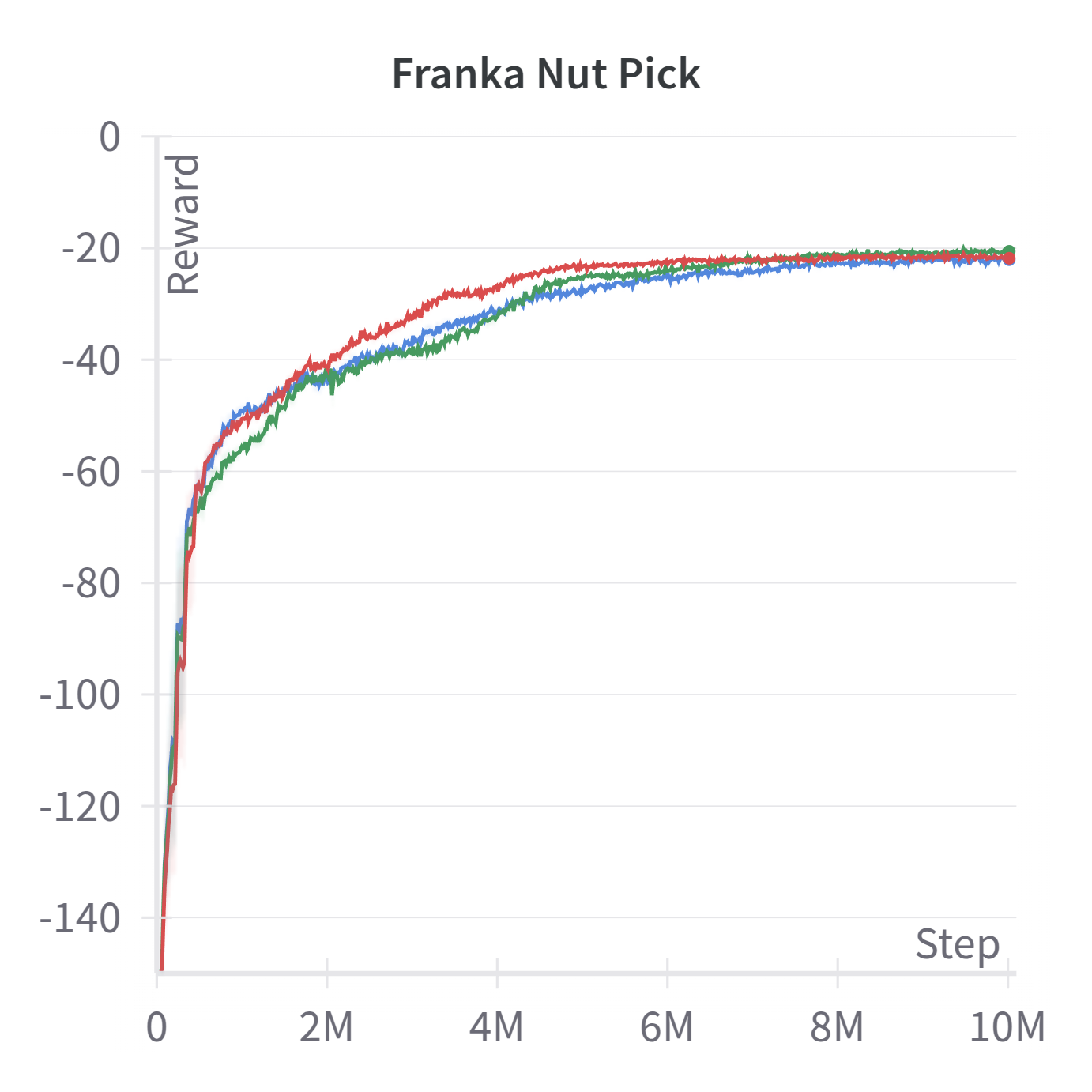}
    \end{subfigure}
    \\
    \begin{subfigure}{0.24\textwidth}
        \centering
        \includegraphics[width=\textwidth]{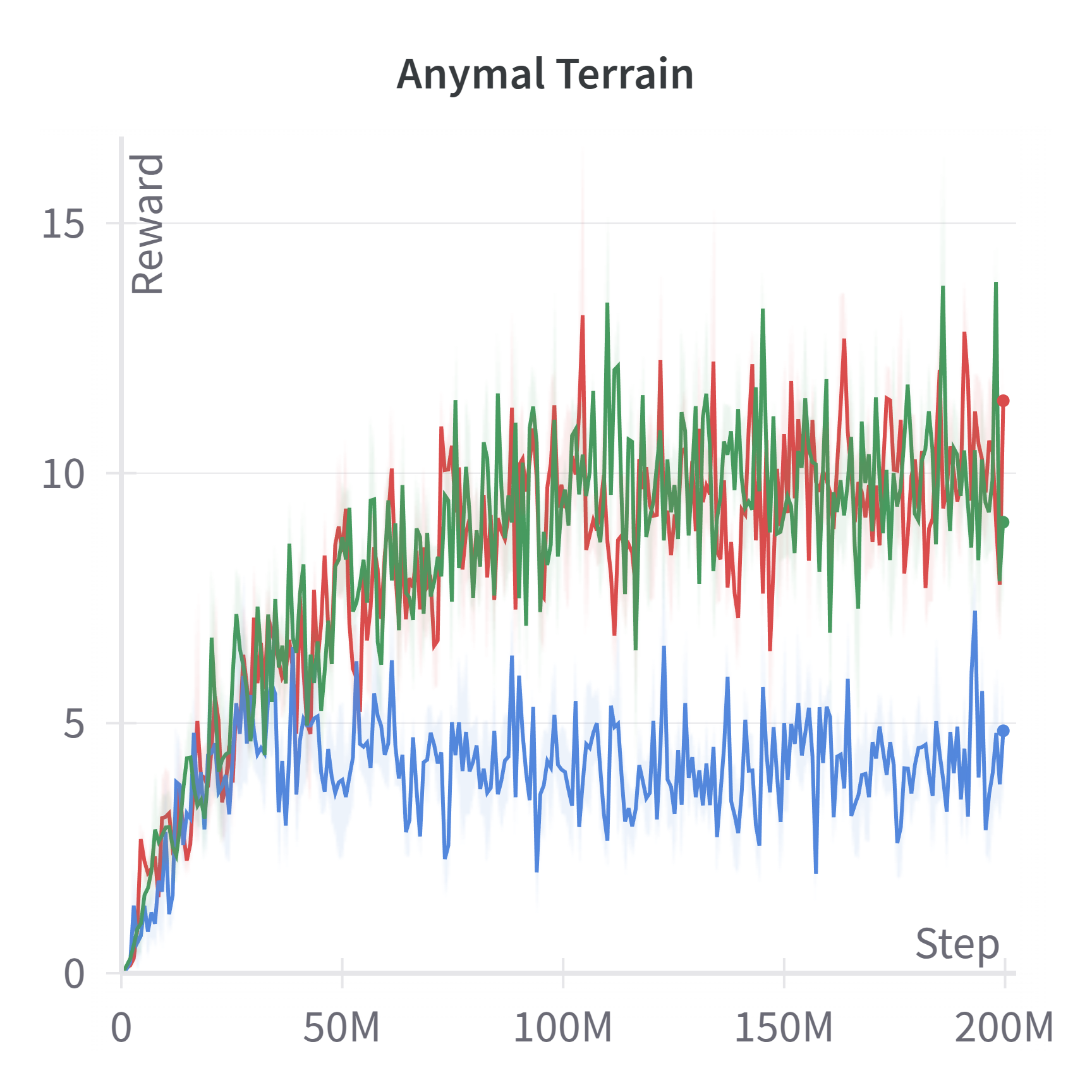}
    \end{subfigure}
    \hfill
    \begin{subfigure}{0.24\textwidth}
        \centering
        \includegraphics[width=\textwidth]{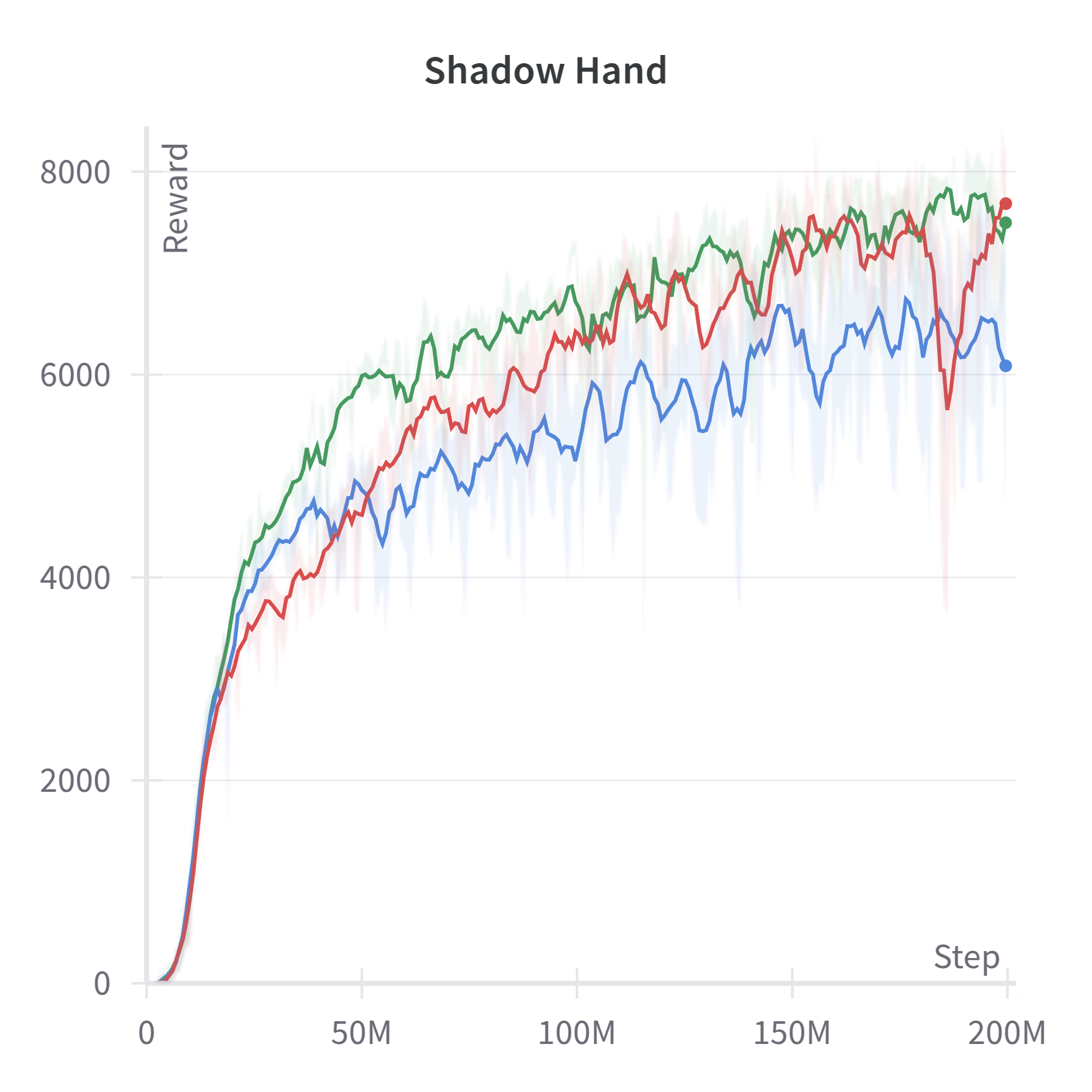}
    \end{subfigure}
    \begin{subfigure}{0.24\textwidth}
        \centering
        \includegraphics[width=\textwidth]{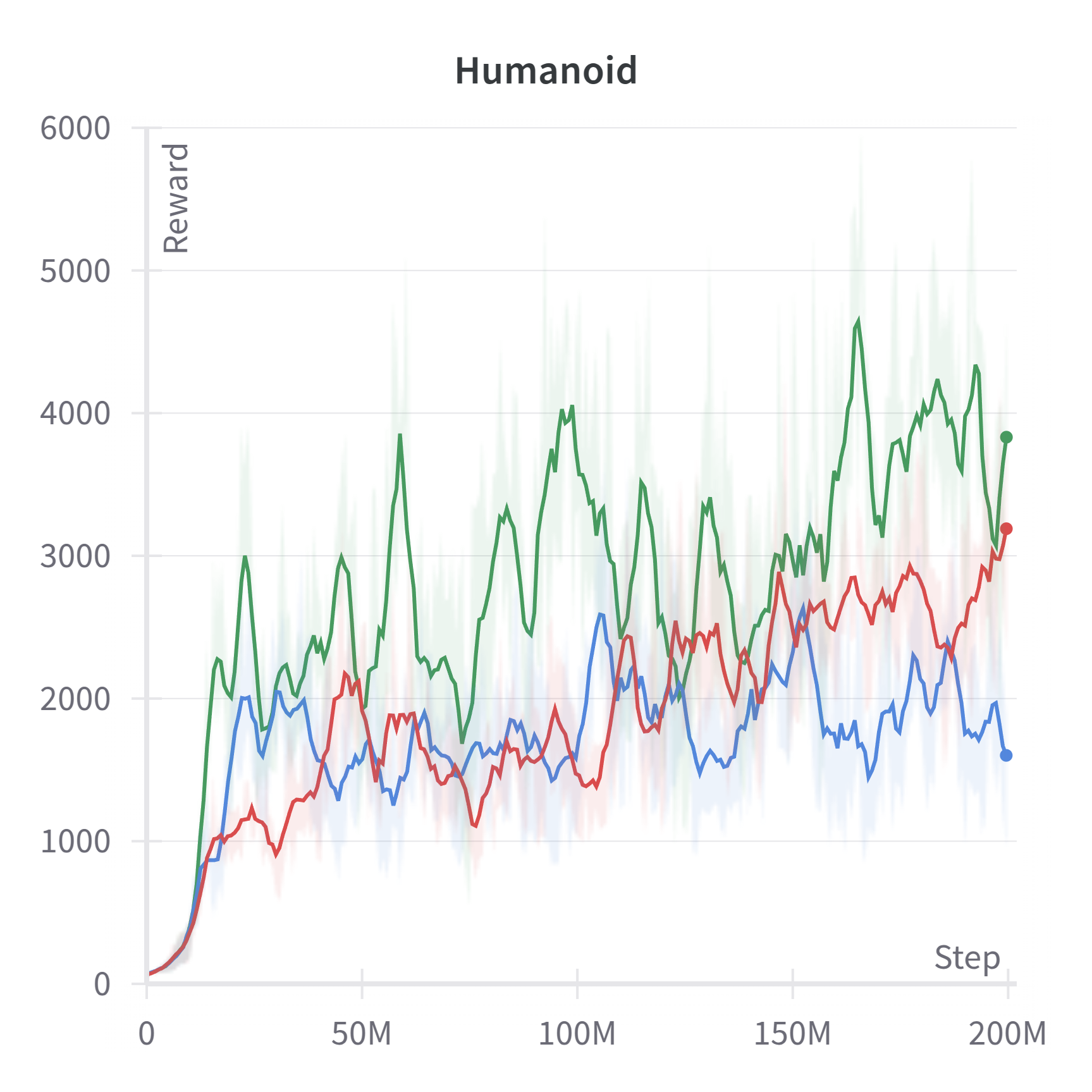}
    \end{subfigure}
    \hfill
    \begin{subfigure}{0.24\textwidth}
        \centering
        \includegraphics[width=\textwidth]{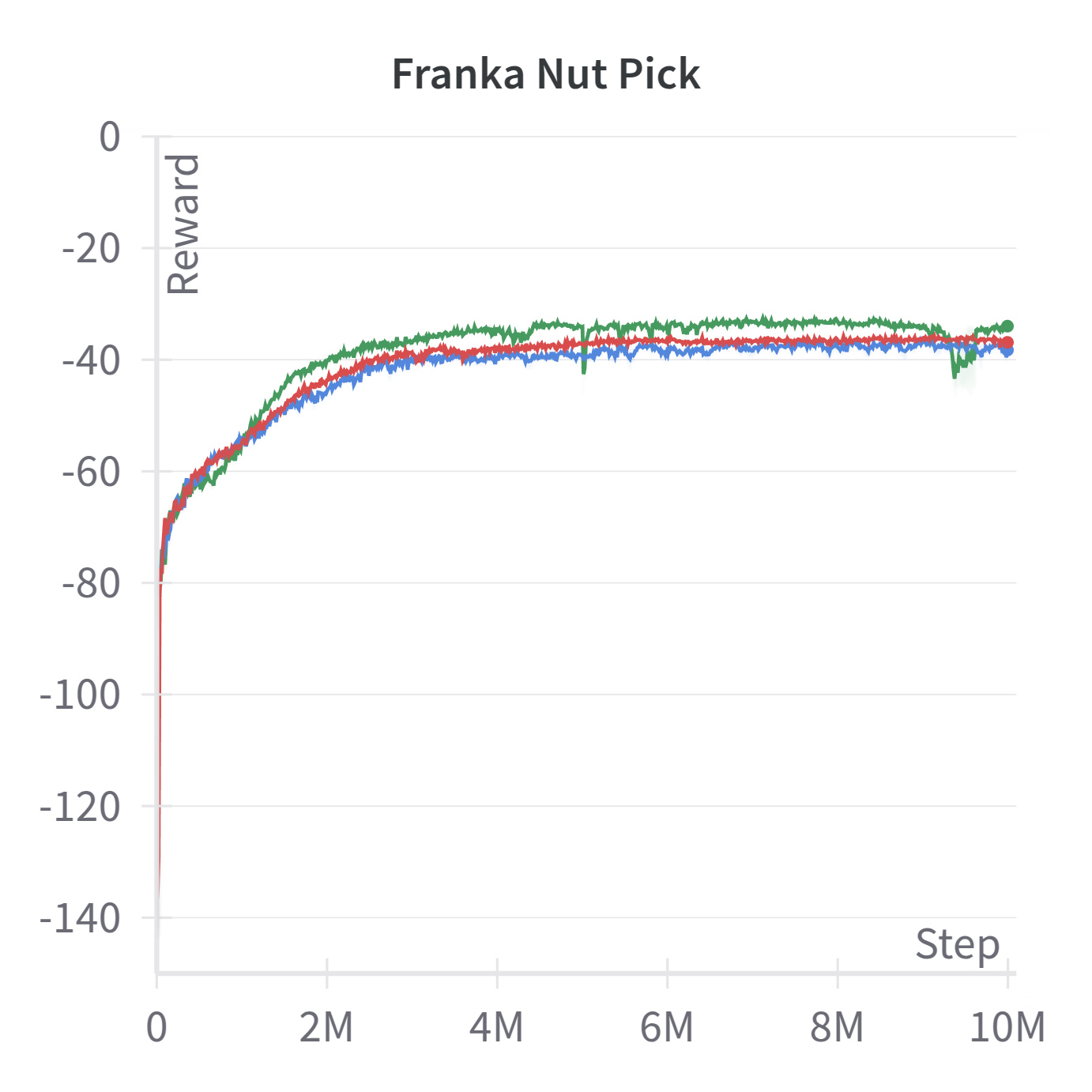}
    \end{subfigure}

    % Caption
    \caption{Comparison of different mutation schemes for PBRL-PPO (top) and PBRL-DDPG (bottom) with $\lvert \mathcal P \rvert = 4$.}
    \label{fig:mutation_comp}
    \vspace*{-1em}
\end{figure*}

\subsubsection{PBRL-DDPG}

In DDPG, exploration noise is added to the output of the deterministic actor.
As mentioned in Section~\ref{sec:rl}, different noise levels are added for different environments uniformly within the range $[\sigma_{\text{min}}, \sigma_{\text{max}}]$.
Both these parameters are crucial in controlling the amount of exploration in DDPG agents.
In PBRL-DDPG, the hyperparameters optimized during training include the minimum and the maximum bounds for noise levels, \ie, $\sigma_{\text{min}}, \sigma_{\text{max}}$, and the LRs of the actor and the critic.
As in PBRL-SAC, the maximum population size in PBRL-DDPG is set to 8 due to the presence of independent replay buffers and GPU memory limitations.
Fig.~\ref{fig:pbrl} shows that PBRL-DDPG achieves significantly better training performance than DDPG on all 4 benchmark tasks.

\subsubsection{Mutation Comparison}\label{sec:mutation-comparison}

Fig.~\ref{fig:mutation_comp} shows the results using 3 different mutation schemes for PBRL-PPO and PBRL-DDPG.
As mentioned in Section~\ref{sec:pbt}, the hyperparameters for under-performing agents are generated either by randomly sampling from their range (see Tables~\ref{tab:hp_ppo}--\ref{tab:hp_off}), by perturbing the parent's values through perturbation factors given in Table~\ref{tab:hp_pbrl}, or through the DexPBT mutation scheme \cite{petrenko2023dexpbt}.
In the latter, the hyperparameters have a $\beta_{\text{mut}} := 0.5$ probability of getting multiplied or divided by a random number sampled from the uniform distribution, $\mu \sim \mathcal U(1.1, 1.5)$.
The results manifest that the perturbed agents either exceed or are on par with the performance of other mutation schemes in 6 out of 8 experiments, with DexPBT mutation scheme performs better with PBRL-DDPG on \textit{Humanoid} and \textit{Franka Nut Pick} tasks, which are less challenging compared to others.
Generally speaking, the perturbation scheme yields less variance (\ie, a more stable learning) compared to DexPBT.

\subsection{Discussion}
\label{sec:disc}

Although PBRL agents outperform non-evolutionary agents in nearly all experiments, the influence of population size across different RL algorithms and tasks does not follow a clear or consistent pattern.
One might intuitively hypothesize that larger, more diverse populations would improve final performance by enabling broader exploration. 
However, our results suggest that increasing the population size does not universally enhance agents training.
This challenges the common assumption that population-based methods inherently benefit from large populations due to their ability of performing a more thorough exploration of the hyperparameter space \cite{jaderberg2017population,parker2020provably}.
Instead, the optimal population size appears to be task- and algorithm-dependent, influenced by factors such as task complexity and the nature of agents interaction.
For simpler tasks, smaller populations may suffice to reach high performance, and additional agents may offer diminishing returns or increase noise during selection.
While our evolutionary mechanism keeps the degree of exploitation by propagating successful hyperparameters, larger populations may dilute this effect by introducing more diverse, potentially suboptimal, behavior patterns.
In contrast, tasks with gradually increasing complexity --- such as \textit{Anymal Terrain}, which uses curriculum learning --- may benefit more from the exploration enabled by larger populations.

Additionally, PBRL performance may lag behind non-evolutionary agents on relatively simpler tasks where optimal hyperparameters are known \textit{a priori}.
This can be noticed on a \textit{Humanoid} task trained with SAC in Fig.~\ref{fig:pbrl}: indeed, baseline policies achieve a higher reward than PBRL-SAC with 4 agents; nevertheless, 8 agents are capable of outperforming the baseline.
Thus, the benefits provided by PBRL will become more apparent for novel tasks lacking known ideal hyperparameter ranges.
In this sense, PBRL can be thought of as an exploratory approach to search for unknown optimal configurations of newly designed tasks.

\subsection{Sim-to-Real Transfer}\label{sec:simtoreal}

For real-world experiments, we replicate the \textit{Franka Nut Pick} task \cite{narang2022factory} by deploying a PBRL-PPO policy, without any real-world adaptation on a physical robot, deploying the actions with PLAI \cite{tang2023industreal}.
The robot detects the nuts utilizing Mask-RCNN \cite{he_mask_2017}, fine-tuned on real-world images captured with a wrist-mounted Intel RealSense D435 camera, using the \texttt{IndustRealLib} codebase \cite{tang2023industreal}.

\begin{table}
\centering
\caption{Simulated environment and real control configuration parameters used in \textit{Franka Nut Pick} during training and deployment, respectively: $\mathcal N$ indicates a Gaussian distribution, while $\pm$ defines a uniform range.}
\label{tab:franka_domain_randomization}
\begin{tabular}{c|c}
\textbf{Parameter} & \textbf{Value} \\ 
\hline
Franka initial position & $\mathcal N([0.0, -0.2, 0.2], [0.2, 0.2, 0.1])$ \\
Franka initial rotation & $\mathcal N([\pi, 0, \pi], [0.3, 0.3, 1])$ \\
Nut initial position & [0.42, 0.27, 0.02] $\pm$ [0.18, 0.13, 0.01] \\
TSI proportional gains & [1000, 1000, 1000, 50, 50, 50] \\
TSI derivative gains & [63.25, 63.25, 63.25, 1.414, 1.414, 1.414] \\
Action scale & 0.0001 \\
\end{tabular}
\end{table}

Compared to the original task \cite{narang2022factory}, we applied the following changes to make the simulated environment as close as possible to the real setup:
\begin{enumerate*}[label=(\roman*)]
    \item employing a Task-Space Impedance (TSI) controller \cite{caccavale_six-dof_1999} instead of an Operational-Space motion Controller (OSC) \cite{khatib_unified_1987} to comply with the actual low-level controller\footnote{The control laws are specified in \cite{narang2022factory} and in reference works \cite{caccavale_six-dof_1999,khatib_unified_1987}};
    \item randomizing the nut's initial position to resemble the actual robot workspace;
    \item changing the observation space to include the 7-DOF joint configuration, the measured end-effector pose, and the estimated nut pose.
\end{enumerate*} The parameters used in the simulated environment and the real controller are reported in Table~\ref{tab:franka_domain_randomization}.

\begin{figure}
    \centering

    % Legend
    \begin{center}
    \footnotesize{
    {\legred}~PPO \hspace{0.3em}
    {\leggreen}~PBRL-PPO~4~agents \par
    {\legblue}~PBRL-PPO~8~agents \hspace{0.3em}
    {\legpurple}~PBRL-PPO~16~agents}
    \end{center}

    % Figure
    \begin{subfigure}[b]{0.49\columnwidth}
        \centering
        \includegraphics[width=\textwidth]{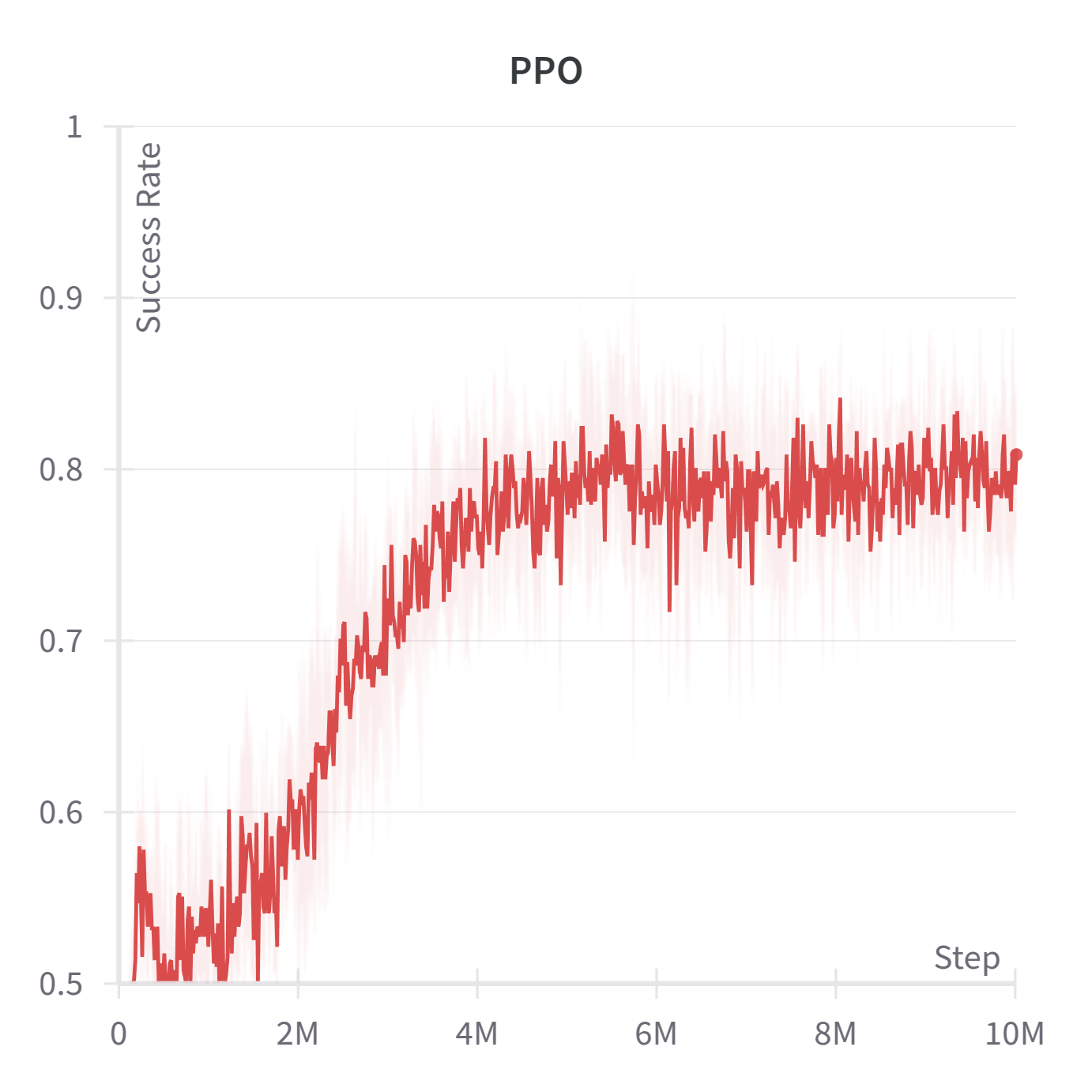}
    \end{subfigure}
    \hfill
    \begin{subfigure}[b]{0.49\columnwidth}
        \centering
        \includegraphics[width=\textwidth]{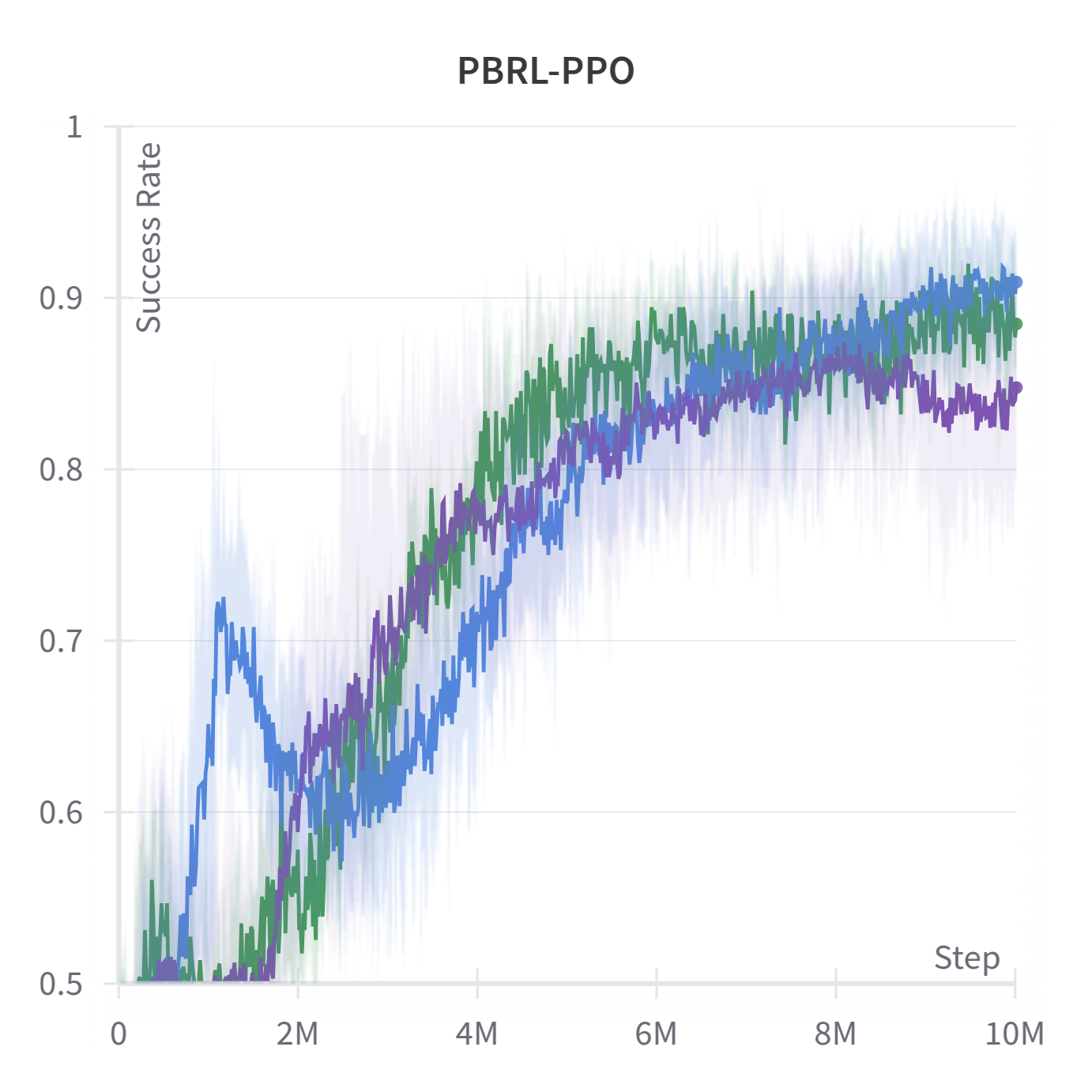}
    \end{subfigure}
    
    % Caption
    \caption{Success rate of \textit{Franka Nut Pick} with PPO baseline and PBRL-PPO in simulation for $\lvert \mathcal P \rvert \in \{4, 8, 16\}$.
    %The shaded area represents the performance between the best and the worst agent in $\mathcal P$, or among 4 different seeds in a non-evolutionary baseline.
    }
    \label{fig:successes}
    \vspace*{-1em}
\end{figure}

During experiments, we deployed the following policies, performing 30 real-world trials of \textit{Franka Nut Pick} task for each policy:
\begin{enumerate*}[label=(\roman*)]
    \item 2 agents from a population of 8, trained with PBRL-PPO, specifically the ``best'' and the ``worst'' agent;
    \item the ``best'' agent trained with baseline PPO.
\end{enumerate*}
With ``best'' and ``worst'' we indicate the agents achieving the highest and lowest \textit{success rate} in simulation, where success is defined as reaching, grasping, and lifting the nut, without losing contact during the lifting phase.
As shown in Fig.~\ref{fig:successes}, PBRL-PPO with $\lvert \mathcal P \rvert = 8$ has the highest success rate.
Remarkably, even the success rate of the worst agent in $\mathcal P$ is higher than that of the best PPO agent ($\approx 90\%$ \vs $\approx84\%$), also highlighting lower variance compared to both baseline PPO and PBRL-PPO with $\lvert \mathcal P \rvert = 16$.

\begin{table}
\centering
\caption{Success rate deploying the best and the worst out of 8 agents trained with PBRL-PPO and the best PPO baseline agent on the \textit{Franka Nut Pick} task with the real robot.}
\label{tab:experiments}
\begin{tabular}{c|c|c|c}
\textbf{Algorithm} & \textbf{Agent} & \textbf{Successful trials} & \textbf{Success rate} \\ 
\hline
PBRL-PPO & Best & 27/30 & 90\% \\
PBRL-PPO & Worst & 21/30 & 70\% \\
PPO & Best & 19/30 & 63.33\%
\end{tabular}
\end{table}

\begin{figure}
    \centering

    % Figure
    \begin{subfigure}[b]{0.32\columnwidth}
        \centering
        \includegraphics[width=\textwidth]{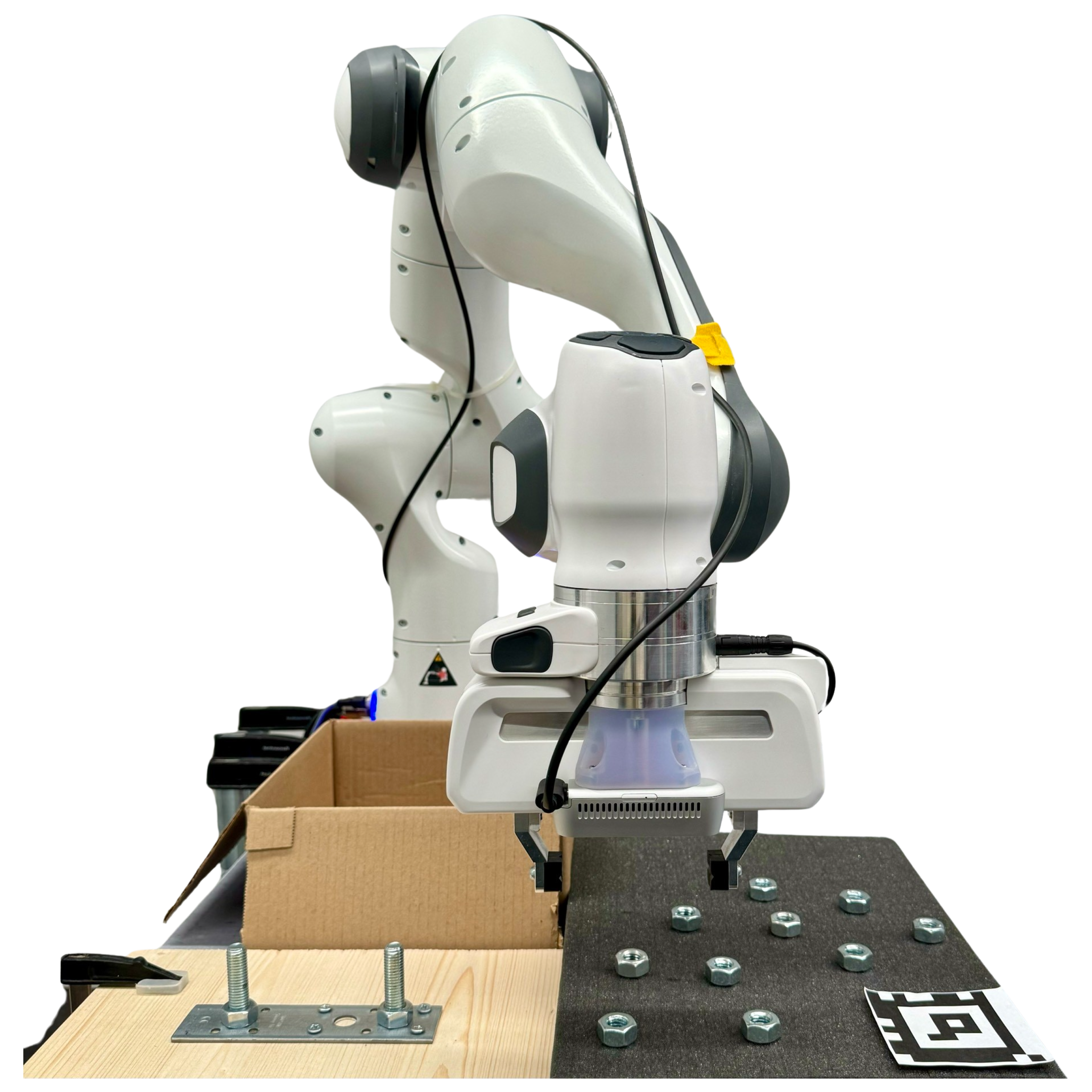}
    \end{subfigure}
    \hfill
    \begin{subfigure}[b]{0.32\columnwidth}
        \centering
        \includegraphics[width=\textwidth]{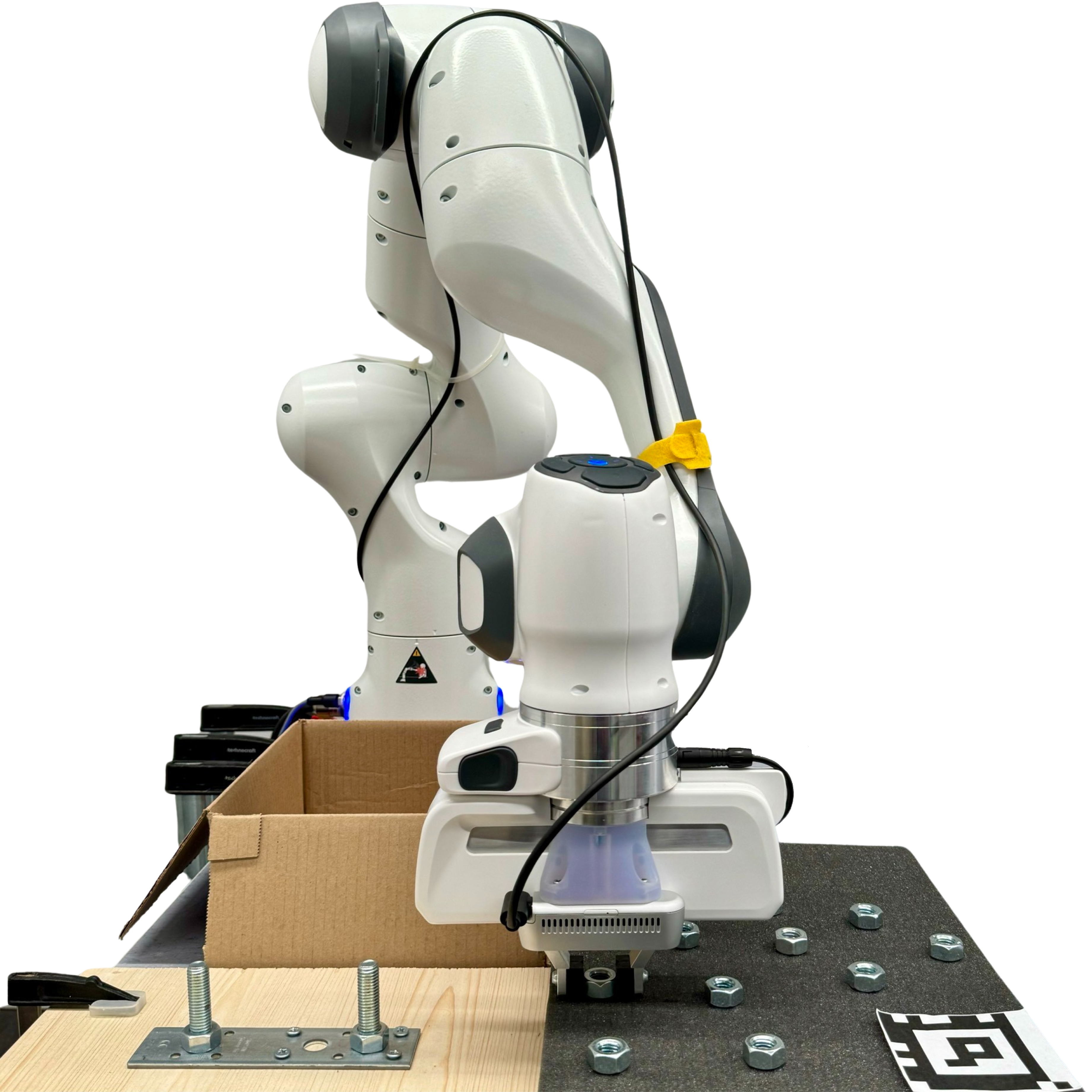}
    \end{subfigure}
    \hfill
    \begin{subfigure}[b]{0.32\columnwidth}
        \centering
        \includegraphics[width=\textwidth]{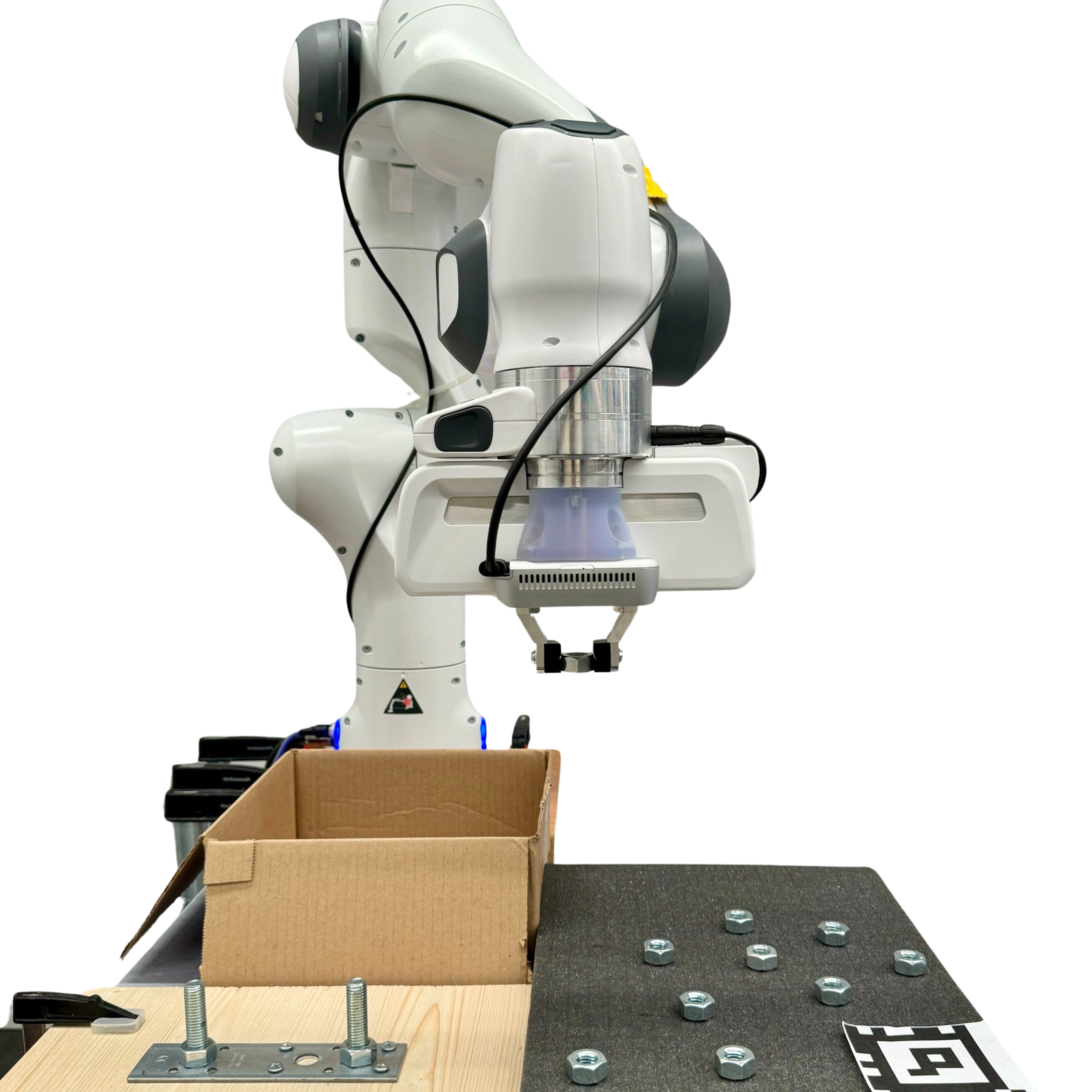}
    \end{subfigure}
    
    % Caption
    \captionsetup{width=\columnwidth}
    \caption{Snapshots of the \textit{Franka Nut Pick} experiment on the real robot: full video at \url{https://sites.google.com/view/pbrl/}.}
    \vspace*{-1em}
    \label{fig:experiment}
\end{figure}

Deploying both PPO and PBRL-PPO agents onto a real robot leads to task completion (shown in Fig.~\ref{fig:experiment}), yet with different success rates, as summarized in Table~\ref{tab:experiments}.
Particularly, both PBRL-PPO agents yield higher success rates than PPO, with the ``best'' agent performing better than the ``worst'' one, indeed confirming the ranking attained in simulation.
This suggests that a better exploration across the agents, favored by PBRL, not only leads to higher rewards, but is also one of the aspects that could lead to a successful real-world transfer \cite{Kim_2025}.
Informally, the best PBRL-PPO agent also exhibited recovery behavior during task execution after perturbation by the human.

\section{Conclusion}

In this paper, a PBRL framework has been employed for training a population of RL agents by making use of high-throughput GPU-accelerated simulation.
The first simulation results of PBRL using on- and off-policy methods are provided on a series of benchmark tasks proposed in \cite{makoviychuk2021isaac} by investigating the effect of population size and different mutation schemes.
The results showed the effectiveness of PBRL in improving final performance by adapting the hyperparameters online.
PBRL agents have been deployed on real hardware for the first time, demonstrating smooth and successful transfer, without any policy adaptation or fine-tuning.
Finally, we released the codebase to train PBRL agents and hope that it will empower researchers to further explore and extend the capabilities of PBRL algorithms on challenging robotic tasks.

Many interesting directions remain for future research.
An immediate extension could be to use a dedicated fitness metric for each population sub-group to prioritize long-term performance \cite{dalibard2021faster}.
This can circumvent a greedy decision process of population-based methods that may lead to undesirable behavior later in the training.

Additionally, it remains to be seen how the PBRL agents perform on contact-rich tasks (e.g., dexterous manipulation, and assembly) in the real world with sim-to-real techniques.
With a number of agents training in parallel environments, the PBRL framework has the potential to solve complex robotic manipulation tasks, making them feasible and computationally tractable, accounting for diverse skills labeling and learning.

\bibliographystyle{IEEEtran}
\bibliography{IEEEabrv,OtherAbbrv,bibliography}

\end{document}